%% file: acl_latex.tex
\pdfoutput=1

\documentclass[11pt]{article}
\usepackage[dvipsnames]{xcolor}

\usepackage{subcaption}
\usepackage{booktabs}
\usepackage{amssymb} 
\usepackage{pifont}

\newcommand\ourenv{\texttt{WhoDunitEnv}}
\newcommand\ourenvsym{\texttt{WhoDunitEnv}-Sym}
\newcommand\ourenvasym{\texttt{WhoDunitEnv}-Asym}
\newcommand\govsim{\texttt{GovSim}}
\newcommand\HumanEval{\texttt{HumanEval}}
\newcommand\LCB{\texttt{LiveCodeBench}}
\newcommand\CG{\texttt{CodeGen}}

\newcommand\gpt{\textsc{GPT-4o}}
\newcommand\llama{\textsc{Llama-3.1-70B}}
\newcommand\qwen{\textsc{Qwen-2.5-72B}}
\newcommand\qwenlarge{\textsc{Qwen-1.5-110B}}
\newcommand\success{Success-Rate}
\newcommand\gamelength{Game-Length}
\newcommand\precision{Precision}
\newcommand\survivalrate{Survival-Rate}
\newcommand\survivaltime{Survival-Time}

\newcommand\efficiency{Efficiency}

\usepackage[final]{acl}

\usepackage{times}
\usepackage{latexsym}

\usepackage[T1]{fontenc}

\usepackage[utf8]{inputenc}

\usepackage{microtype}

\usepackage{inconsolata}

\usepackage{graphicx}
\usepackage{amsmath}
\usepackage{amsfonts}
\usepackage{multirow}

\usepackage[hang,flushmargin]{footmisc}
\definecolor{light_blue}{HTML}{eec9f8}
\usepackage[most]{tcolorbox}
\tcbset{on line,
        boxsep=1pt, left=0pt,right=0pt,top=0pt,bottom=0pt,
        colframe=white,colback=light_blue,
        highlight math style={enhanced}
        }
\title{Preventing Rogue Agents Improves Multi-Agent Collaboration}

\author{
 \vspace{5px}
Ohav Barbi ~~~~~~~~~ Ori Yoran ~~~~~~~~~ Mor Geva \\ \vspace{3px}
Blavatnik School of Computer Science and AI, Tel Aviv University\\
\small{\texttt{\{ohavbarbi@mail,ori.yoran.cs,morgeva@tauex\}.tau.ac.il}}
}

\begin{document}
\maketitle

\begin{abstract}

Multi-agent systems, where specialized agents collaborate to solve a shared task hold great potential, from increased modularity to simulating complex environments. However, they also have a major caveat -- a single agent can cause the entire system to fail. Consider a simple game where the knowledge to solve the task is distributed between agents, which share information in a communication channel. At each round, any of the agents can terminate the game and make the final prediction, even if they are uncertain about the outcome of their action. Detection of such rogue agents \textit{before they act} may prevent the system's failure. 
In this work, we propose to \emph{monitor} agents during action prediction and \emph{intervene} when a future error is likely to occur. To test our approach, we introduce \ourenv{}, a multi-agent collaboration environment that allows modular control over task complexity and communication structure. Experiments on \ourenv{}, code generation tasks and the \govsim{} environment for resource sustainability show that our approach leads to substantial performance gains up to 17.4\%, 2.5\% and 20\%, respectively. Thorough analysis shows that our monitors successfully identify critical points of agent confusion and our interventions effectively stop agent errors from propagating. We release \ourenv{} and our code for future studies on multi-agent collaboration at 
\url{https://github.com/Ohav/rogue-agents}.

\end{abstract}

\input{latex/sections/01_intro}

\input{latex/sections/02_03_setup_env}

\input{latex/sections/04_experimental_settings}

\input{latex/sections/05_results}
\input{latex/sections/06_related_work}
\input{latex/sections/07_conclusion}

\input{latex/sections/limitations}

\section*{Acknowledgments}
We thank Amit Elhelo for valuable feedback.
This research was supported in part by AMD's AI \& HPC Fund, the Google PhD Fellowship program, Len Blavatnik and the Blavatnik Family foundation.
Figures~\ref{fig:intro},~\ref{fig:asym_env},~\ref{fig:sym_env}, and~\ref{fig:examples} use images by Rank Sol on IconScout and from Freepik. The authors used AI models, specifically ChatGPT and Gemini, for implementing helper functions.

\bibliography{custom}

\appendix
\input{latex/sections/appendix}

\end{document}

%% file: latex/sections/01_intro.tex
\section{Introduction}
\label{sec:intro}

Recently there has been a growing interest in leveraging large language models (LLMs) to solve complex multi-agent tasks \cite{guo2024largelanguagemodelbased,tran2025multiagentcollaborationmechanismssurvey}, such as improving factuality and reasoning, \cite{du2023improvingfactualityreasoninglanguage, liang-etal-2024-encouraging} simulating software development teams \cite{hong2024metagpt, qian-etal-2024-chatdev, liu2024a}, and human interactions \cite{10.1145/3586183.3606763}. A key requirement for succeeding at such cooperative tasks is the agent's ability to communicate effectively.

\begin{figure}[t]
\setlength\belowcaptionskip{-10pt}
\centering
  \includegraphics[width=\columnwidth]{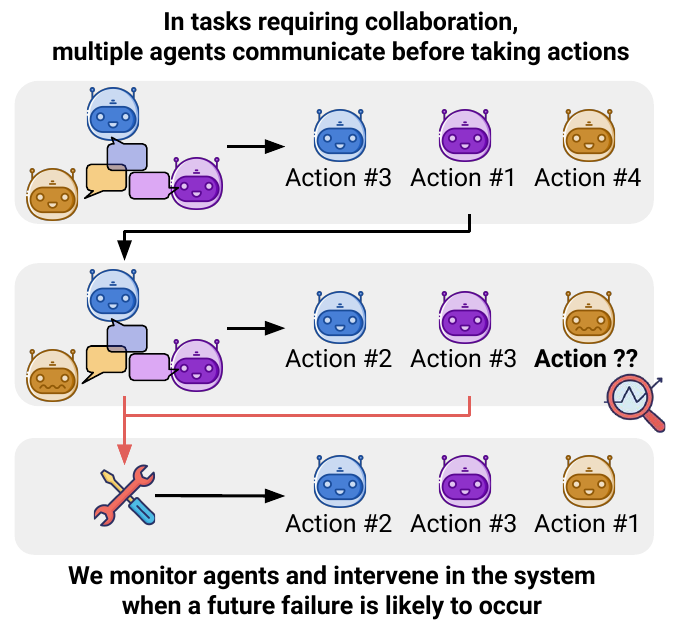}
  \caption{\textbf{An overview of our approach}. We propose to improve multi-agent collaboration by monitoring agent communication and applying interventions to the environment in case rogue agents are detected.}
  \label{fig:intro}
\end{figure}

However, establishing effective communication is a major challenge for current LLMs, which often ignore critical information in the communication \cite{liu-etal-2024-lost, levy-etal-2024-task}, become distracted by irrelevant messages \cite{shi2023large, amayuelas-etal-2024-multiagent}, and introduce hallucinations \cite{xiao-wang-2021-hallucination, zhang2023sirenssongaiocean, jiang2024surveylargelanguagemodel, lin2024maoframeworkprocessmodel, Huang_2025}.
Such failures introduce noise to the  communication, which can be amplified as LLM responses are generated for many steps \cite{zhang2024how, ivgi2024from, yoran-etal-2024-assistantbench, jimenez2024swebench}. A single valid agent can be consequently affected by these issues and go \textit{rogue}. This \textit{rogue agent} could then cause the entire system to fail.

To improve multi-agent collaboration, recent works have proposed introducing changes to the communication protocol \cite{guo2024embodiedllmagentslearn, mehta-etal-2024-improving}, augmenting the communication with agent beliefs \cite{li-etal-2023-theory}, and adding specific modules for reasoning and grounding \cite{agashe2023evaluating, hong2024metagpt}.
While these approaches often yield improvements in the overall task performance, they do not prevent cases where one \textit{rogue agent} introduces noise to the communication and drags the system into failure \cite{darcy2024margmultiagentreviewgeneration, hong2024metagpt}.

In this work, we propose to improve multi-agent collaboration through live monitoring and interventions in agent communication. Specifically, we hypothesize that many failures can be detected and prevented early on --- before the agent becomes non-functional and the task fails --- based on the state of the agent and how it observes the environment (see Fig.~\ref{fig:intro} for an illustration). 
This approach is widely used in industrial systems, such as intrusion detection software and critical manufacturing processes \cite{STAVROPOULOS2013421, 10.1145/3411763.3451774, manpaper, Woods02092023, cyber}, and is also observed in natural systems such as the human immune system, which monitors faulty behavior and prevents infections from spreading throughout the body.

To test this hypothesis, we present a monitor and intervention framework (\S\ref{sec:method}). To monitor an agent's state, we employ simple classifiers that detect rogue agents based on intrinsic signals of uncertainty, such as the prediction's entropy during action selection. This monitor then triggers an intervention mechanism, which prevents the communication from being distorted and the agents from failing at the task. 
Specifically, we experiment with simple interventions that roll back certain parts of the communication or the entire environment.

To evaluate our approach, we create a modular collaborative environment, called \ourenv{}, which allows analyzing different agent communication protocols at varying task-complexity levels (\S\ref{sec:env}). \ourenv{} is inspired by the well-known game of \emph{Guess Who},\footnote{\url{https://en.wikipedia.org/wiki/Guess_Who}} where players (here agents) collaborate to identify a culprit among a group of suspects based on their appearance properties. 

We conduct experiments (\S\ref{sec:experimental_setting}) on two variants of \ourenv{}, a code generation environment \CG{}, and the \govsim{} resource sharing environment \cite{piatti2024cooperate}. These environments cover symmetric and asymmetric agents, structured and free-form communication protocols, and varying levels of task complexities. 

Our results (\S\ref{sec:results}) show that, across all settings, monitoring rogue agents leads to substantial improvements in multi-agent collaboration. Specifically, we observe gains of up to 20\%.
Finally, we conduct ablations and qualitative analysis (\S\ref{sec:analysis}), showing that strong monitors and interventions are needed to improve performance, and that our monitors effectively identify a range of \emph{rogue} agent behaviors, including hallucination of information in the communication channel.

To summarize, our work makes the following contributions: (a) we propose the notion of live monitoring and interventions to prevent failures due to rogue agents in multi-agent systems, (b) we introduce \ourenv{}, a modular environment for studying collaborative multi-agent systems where agents are either symmetrical/asymmetrical, (c) we show that our approach leads to substantial performance gains on \ourenv{}, \CG{}, and \govsim{} across models and complexity levels, and (d) we show that improvements are due to strong monitors and interventions.

%% file: latex/sections/02_03_setup_env.tex
\section{Monitoring and Intervening in Multi-Agent Systems}
\label{sec:method}

In multi-agent systems, agents collaborate in order to solve a task or use shared resources.
The system comprises of agents $G=\{g_i\}$, a communication channel $C$ that stores messages from the agents, and a shared task $T$. Agents possess knowledge $K_i$ and perform actions $A_i$. 
For example, agents that simulate human behavior have different memories based on their experiences \cite{10.1145/3586183.3606763}, and different personas in a software development team have different actions, such as designing the code, programming, or writing tests \cite{qian-etal-2024-chatdev}.

Let $C_j$ be the communication channel at the $j$-th turn, an agent $g_i$ chooses their next action using a probability distribution over actions conditioned on their knowledge and the shared information: 
\[P_{a \in A_i}(a, j, g_{i}, T)=P(\texttt{Action}=a \,|\, K_i, C_j, T)\]
Similarly, agents share information in the channel with a distribution over knowledge pieces:
\[P_{k \in K_i}(k, j, g_{i}, T)=P(\texttt{Share}=k \,|\, K_i, C_{j-1}, T)\]

We propose to perform \emph{live}, mid-run interventions to prevent single agents from causing a system-wide failure (Fig.~\ref{fig:intro}). Our approach consists of monitoring agent action predictions to detect rogue agents, and intervening in the environment when a rogue agent is detected.

We view \textit{monitoring} as a function that estimates the probability of succeeding at the task at every turn, based on the agent's probability distribution over actions.
Namely, given $P_{A_i}$ for agent $g_i$ at turn $j$, we wish to estimate $P(\texttt{success} \,|\, P_{A_i}, j, g_i, T)$. 

If a future failure is likely, namely, $P(\texttt{success} \,|\, P_{A_i}, j, g_i, T) < \tau$ for some threshold $\tau$, we intervene to provide agents with an opportunity to reach a better state. An \textit{intervention} is a causal operation that modifies the current state of the environment based on its current state and the monitoring output. For example, the intervention could revert the communication or augment it with additional content.

\paragraph{Monitoring agent uncertainty to predict failures}
Inspired by prior work on agents in Reinforcement Learning and NLP environments \cite{acharya2022uncertaintyquantificationcompetencyassessment, liu-etal-2024-uncertainty, doi:10.1126/sciadv.adk1256, debunc}, we predict task success based on agent uncertainty. Namely, if the agent is ``confused'' in their action selection, they are likely to introduce noise which could fail the whole system. 

Let $\mathbf{p}_i$ be the output probability distribution vector at position $i$.\footnote{For proprietary models, where we cannot access to the full probability distribution, we approximate $\mathbf{p}_r$ with the top $k$ tokens, setting $k=10$.} We consider all the positions in the agent's generation that hold important information for its action selection. These include both the final action selection and the preceding thoughts, and differs between different environments (exact definition for each available in section \S\ref{sec:experimental_setting}). To estimate the agent's uncertainty, we compute the entropy, varentropy, and kurtosis of $\mathbf{p}_i$ (see exact definitions in \S\ref{sec:stat_measures}) and take their maximum values over all selected positions, in addition to the current turn count, as features.
We use the features to fit a simple polynomial ridge classifier with the goal of estimating the success probability $f: \mathbb{R}^m \rightarrow [0,1]$, where $m\leq4$ is the number of features used. This is done using boolean labels from training games at each intermediate game turn. During test time, these features are collected at every turn and are fed into the monitor, which outputs the probability of success. Further details are in \S\ref{sec:experimental_setting}.

\paragraph{Live interventions to prevent system failures}
When the monitor estimate is lower than $\tau$, the monitor triggers an intervention. 
Notably, environments can have both \textit{reversible} and \textit{irreversible}.
\textit{Irreversible} actions include using a shared resource or committing to a solution, while \textit{reversible} actions consist of only sharing information. 
When intervening, we undo \textit{reversible} actions until the last \textit{irreversible} action, to preserve the realism of our method and provide agents with another opportunity to collaborate.

Next, we introduce a multi-agent collaboration environment and evaluate our approach.

\section{\ourenv{}: An Environment for Multi-Agent Collaboration}
\label{sec:env}

\ourenv{} is a modular multi-agent environment, where agents act as detectives working together to point out a culprit out of a suspect lineup. A game is comprised of $N$ suspects, each with a unique set of attribute-value pairs that are randomly assigned from a predefined set. Attributes include clothing (e.g., a green shirt), accessories (e.g., a silver watch), and personality traits (e.g., mood). One suspect is randomly chosen as the culprit. Each agent receives partial information $K_i$ about the suspects or the culprit, and must collaborate to find the culprit and accuse them. Turns move in a round robin fashion, and the game ends once either an agent accuses a suspect or a turn limit is reached.
Actions are tuples $(a,t)$, consisting of a prime action $a\in A$ and a target $t$ to which $a$ is applied.
We provide two variants of the environment, asymmetric and symmetric, that differ by the action set $A$ and information available to each agent $K_i$. See \S\ref{app:prompts} for prompts and \S\ref{app:full_example} for an example.

\begin{figure}[t]
\setlength\belowcaptionskip{-10pt}
    \centering
    \includegraphics[width=0.8\linewidth]{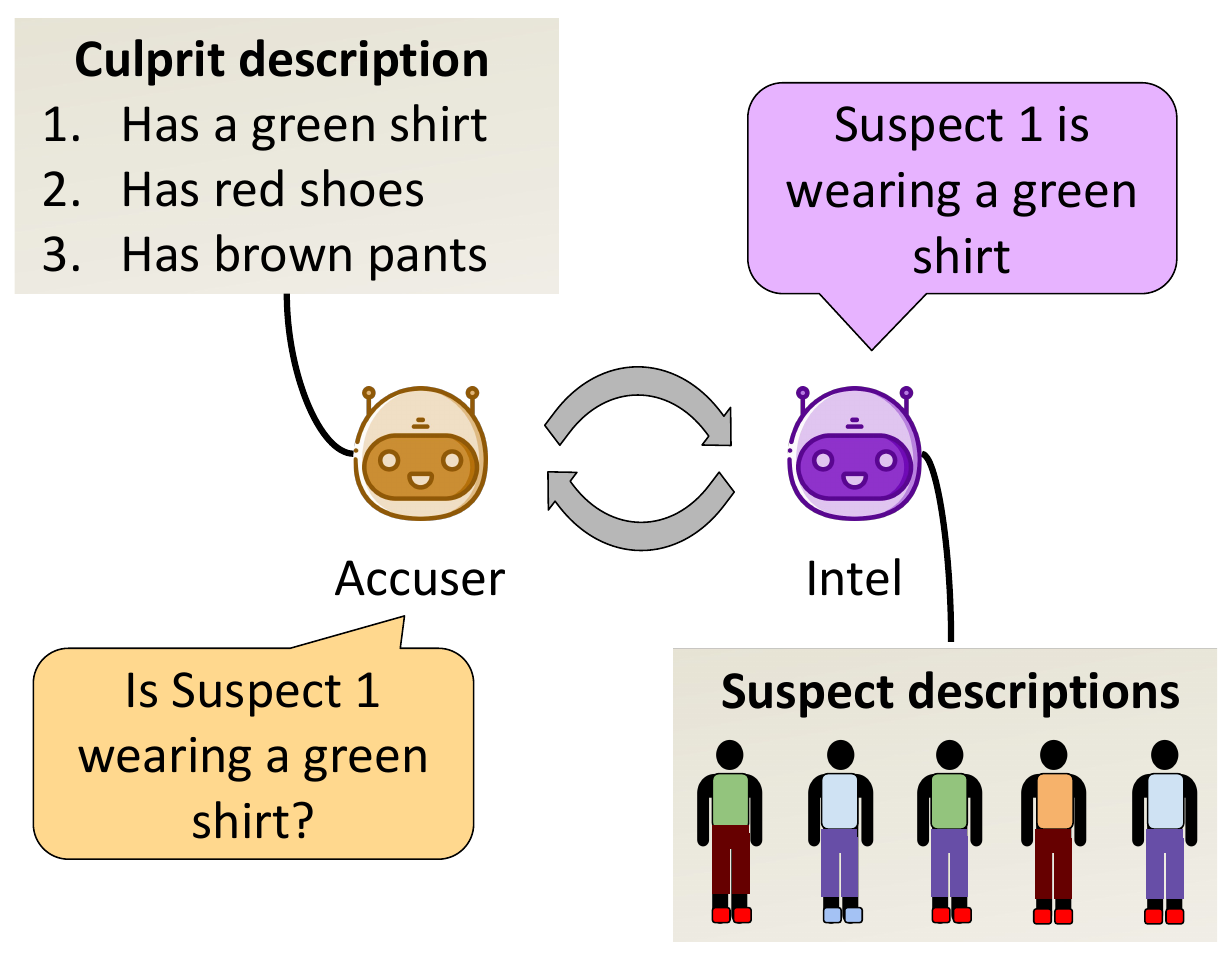}
    \caption{\textbf{An illustration of \ourenvasym{}} Accuser and Intel collaborate to identify the culprit from a lineup of suspects. Accuser, knowing the culprit's identity, can query and accuse. Intel chooses what and how much information to provide about the suspects.}
    \label{fig:asym_env}
\end{figure}

\paragraph{\ourenvasym{}} (Fig.~\ref{fig:asym_env})
This variant consists of exactly two agents -- Accuser and Intel. $K_{accuser}$ contains the exact description of the culprit, but does not contain any information about the suspects. $K_{intel}$ is the complete description of all suspects, without any indication of the culprit. The set of actions available to Accuser is $A_{accuser}=\{\texttt{request-specific}, \texttt{request-broad}, \texttt{accuse}\}$, which allows it to request information about a specific attribute of a suspect, request broad information from Intel for no specific suspect or attribute, and accuse a suspect, respectively. The set of actions available to Intel is $A_{intel}=\{\texttt{respond}, \texttt{respond-broad}\}$ which allows it to respond to Accuser's request with a yes/no answer or return a broad message. When returning a broad message, the agent decides on an specific attribute value, such as ``green hat'', then lists all the suspects that have this property. Thus, Intel can choose to provide different (broader) information than requested by Accuser.

\begin{figure}[t]
\setlength\belowcaptionskip{-10pt}
    \centering
    \includegraphics[width=0.91\linewidth]{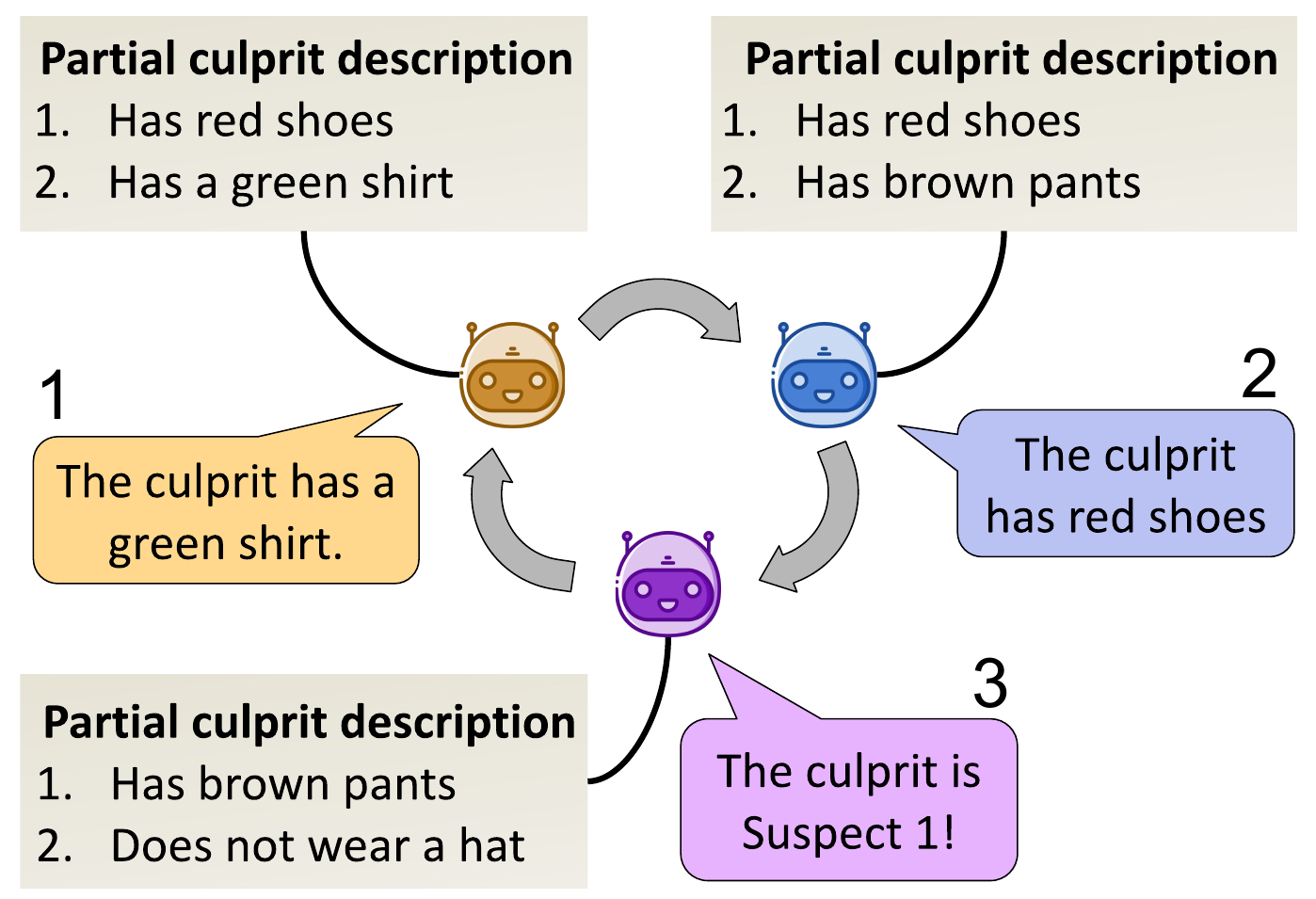}
    \caption{\textbf{An illustration of \ourenvsym{}}. Agents are tasked with identifying the culprit among a lineup of suspects by sharing information they posses. Information about the culprit is equally spread across the agents and all agents can accuse a suspect.}
    \label{fig:sym_env}
\end{figure}

\paragraph{\ourenvsym{}} (Fig.~\ref{fig:sym_env})
In the previous environment, the agents are asymmetric in terms of the actions they can perform. Here, we propose a variant where all agents are equal in their actions, but different in the knowledge they posses. Agents start with full knowledge of all the suspects and their attributes, but each agent is given a different set of starting facts about the culprit $K_i=\{f_i^{(1)}, f_i^{(2)}, f_i^{(3)}\}$, where every fact is an attribute value. In each turn, an agent chooses an action $a\in \{\texttt{share}, \texttt{accuse}, \texttt{skip}\}$. For $a=\texttt{share}$, the agent selects a fact from $K_i$ and outputs it in a message to the rest of the group. For $a=\texttt{accuse}$, the agent decides a suspect to accuse of being the culprit and with that ends the game. With $a=\texttt{skip}$, the game simply moves to the next agent, spending the turn.

\paragraph{Task complexity}
\ourenv{} has different levers for increasing environment complexity, which allows adjusting it for evaluation of agents with different capabilities. 
These levers include: (a) \textit{suspect count}: changing the suspect count can change how long the starting context is and the probability of having two very similar suspects, (b) \textit{attribute count}: changing the number of attributes each suspect has can create more specific suspects that are harder to set apart, and (c) \textit{turn count}: the game is set at a time limit, which affects the behavior of agents. By limiting their time, we force agents to use their available information better.

%% file: latex/sections/04_experimental_settings.tex
\section{Experiments}
\label{sec:experimental_setting}

We conduct experiments on \ourenv{} and two other environments. For each environment, we use train and validation sets to train our monitors and a separate test set to evaluate them, as detailed below.


\subsection{Environments}

\paragraph{\ourenv{}} We wish to have a challenging yet feasible environment. To this end, we set the number of suspects, turn limit, and number of attributes in the asymmetric variant to 10, 31, and 11, respectively. For the symmetric variant we set them all to 20.
Attributes include clothing and personality-related features, while each attribute has 2-3 possible values (see the full list in \S\ref{app:attributes}).

In our experiments, we observed that agents often struggle to perform the task when it is described with the words ``accuse'', ``suspect'' and ``culprit'', potentially due to alignment procedures. Therefore, we rephrased the task with the words ``award'', ``character'' and ``winner'', instead. This does not affect the task itself but only how it is presented to the agents.
We prompt agents to generate thoughts before predicting an action with ReAct \cite{yao2023react} (see \S\ref{app:prompts} for the exact prompts). For monitoring, we use all the positions where the agent generated a suspect ID. We treat information sharing as a \emph{reversible} and accusations as \emph{irreversible}, so an intervention consists of resetting the entire game, as long as an accusation was not made. 

We use 210, 90 and 180 environment instances as training, validation and test sets, respectively. Sets are separated by suspect descriptions and the culprit choice.
Performance is measured by the percentage of games that end in identifying the culprit, termed \success{}. Additionally, we measure \precision{}, i.e. \success{} when a character was accused, and \gamelength{} for the average number of dialog turns, including interventions.

\paragraph{\CG{}}
Code generation tasks are a popular practical application for LLMs.\cite{jimenez2024swebench, Xia_2022, min2024beyond}.Typically, agents are given a task description and a starting code snippet, and tasked to produce Python code that solves the task \cite{chen2021evaluatinglargelanguagemodels,jain2024livecodebenchholisticcontaminationfree}.
Inspired by debate frameworks \cite{du2023improvingfactualityreasoninglanguage, liu2024dynamicllmpoweredagentnetwork} we leverage a multi-agent debate system for code generation, \CG{}. We assign 4 different coding roles, a judge, and a unit test writer (see \S\ref{app:codegen}). The first round consists of each coding agent $g_i$ outputting their solution $s_i^0$. Following rounds $j$ consist of the judge and tester evaluating previous solutions $S^{j-1}$ and providing feedback $F^j$ and a list of tests $T^j$. Coding agents then receive $(S^{j-1},\,F^j,\,T^j)$ and attempt solving the task again, also outputting their measurement of how helpful $F^j$ and $T^j$ were.
Passing information through the judge is a basic form of communication, which we monitor. We treat code generation as irreversible, with the intervention consisting of the judge and tester writing new $(F^j, T^j)$.

We use the \HumanEval{} benchmark \cite{chen2021evaluatinglargelanguagemodels} to train and validate our monitor with a 70-30 split. Final test evaluation is done on 279 problems from \LCB{} \cite{jain2024livecodebenchholisticcontaminationfree} that were collected between August 2024 and February 2025, to assess the generalization between similar tasks. We report Pass@1, which measures the percentage of problems correctly solved by the environment.

\paragraph{\govsim{}} \cite{piatti2024cooperate} is a collaborative environment for resource sustainability, where agents share a renewable starting resource $R_0$ that they harvest to maximize their gains. At every round, agents harvest the resource and then discuss and decide their plans for future rounds. At the beginning of each round $j$, the remaining resources $R^*_j$ are doubled up to the original maximum: $R_{j+1} = \min(2R^*_j, R_0)$, encouraging agents to be efficient while ensuring sustainability. The discussion stage allows free communication between agents.
In our experiments, we focus on the fishing task. The set of actions at turn $j$ is defined as the possible amounts of resources to harvest, that is $a\in[0, R_j]$.
To calculate features for our monitor, we consider all the position corresponding to mentions of resource amounts. Since actions consume the shared resource, they are \emph{irreversible}, and the intervention consists of resetting only the previous communication round and allowing the agents to discuss again, without any knowledge of the reset or any reflection made after the last discussion. 

We follow \citet{piatti2024cooperate} and report \survivaltime{}, \survivalrate{} and \efficiency{}.
\survivaltime{} measures the number of rounds in which the shared resource remains above a minimal threshold $\gamma$.
\survivalrate{} is a binary metric indicating whether the \survivaltime{} is above a maximal round threshold $m$. \efficiency{} measures how well agents consume the shared resource, i.e.,
\[
\text{Efficiency} = 1 - \frac{\max(0, c-\sum_j^m\sum_i^n \text{r}(g_i,j))}{c}
\]
where $\text{r}(g_i,j)$ is the amount of fish consumed by agent $g_i$ at round $j$ and $c=\frac{m\cdot R_0}{2}$.
Following \citet{piatti2024cooperate}, we set $\gamma=5, m=12$. Training, validation and testing sets consist of 26, 14 and 20 instances, respectively, and differ by their starting resource $R_0$.\footnote{We extend the evaluation by \citet{piatti2024cooperate} from 5 to 20 games to obtain a better performance estimate.} Exact $R_0$ values are given in \S\ref{app:datasplits}.

\subsection{Monitoring \& Intervention}
For every environment and agent type, we train classifiers with different feature combinations, polynomial degrees $d\in[1,5]$ and monitoring threshold $\tau\in[0,1]$. From these classifiers, we choose the one that best performs on the validation set (see details in \S\ref{app:classifiers} and Tab.~\ref{tab:classifiers_used}). This results in a simple monitor that estimates $P(\texttt{success})$ at every turn.
For the interventions, we define a cap on the number of triggers, setting it to either one or two in \ourenv{} and one in \govsim{} and \CG{}. Agents with different roles  exhibit different behavior and so can each trigger up to the cap.

\subsection{Models}
For \ourenv{}, we experiment with two strong open-weight models \llama{} \cite{grattafiori2024llama3herdmodels} and \qwen{} \cite{qwen2025qwen25technicalreport}, and one proprietary model -- \gpt{} \cite{openai2024gpt4technicalreport}. We report average scores and standard error over four runs with \textsc{Llama} and \textsc{Qwen} and three runs with \gpt{}.

For \CG{}, we use \llama{}.
In \govsim{}, the performance of \textsc{Llama-3-70B} and \textsc{Qwen-1.5-72B} is near-zero \cite{piatti2024cooperate},\footnote{We observed similar results with the newer \llama{} and \qwen{}.} which leaves no positive examples for monitor training (we discuss applicability of our approach in the Limitations section). Thus, we use the stronger \qwenlarge{} \cite{bai2023qwentechnicalreport} and \gpt{} (see more details in \S\ref{app:models}).

%% file: latex/sections/05_results.tex
\begin{figure}[t]
    \centering
        \includegraphics[width=\linewidth]{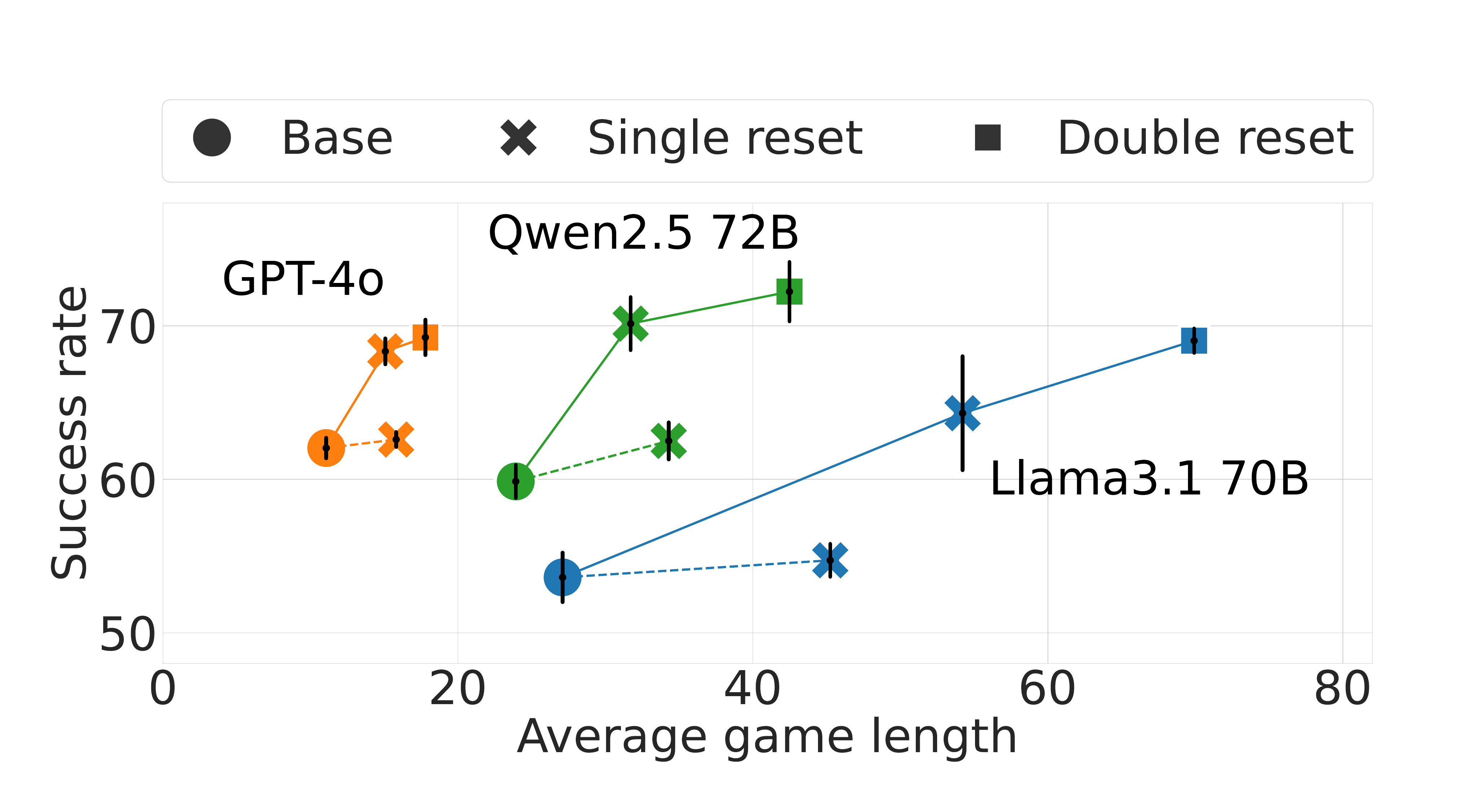}
        \caption{\textbf{Main results for \ourenvasym{}}. Live interventions lead to significant \success{} gains for three strong LLMs of up to 10 points, outperforming the random baseline (a cross connected by a dashed line). Resetting twice leads to additional, yet diminishing gains. Black lines indicate standard error.}
        \label{fig:asymmetric-env-results}
\end{figure}

\section{Results}
\label{sec:results}

\textbf{We find that monitoring and interventions consistently improve system performance across all environments and models, leading to substantial gains of up to 17.4\% in \ourenv{}, up to 20.0\% in \govsim{} and up to 2.5\% in \CG{}.}

\begin{figure}[t]
    \centering
        \includegraphics[width=\linewidth]{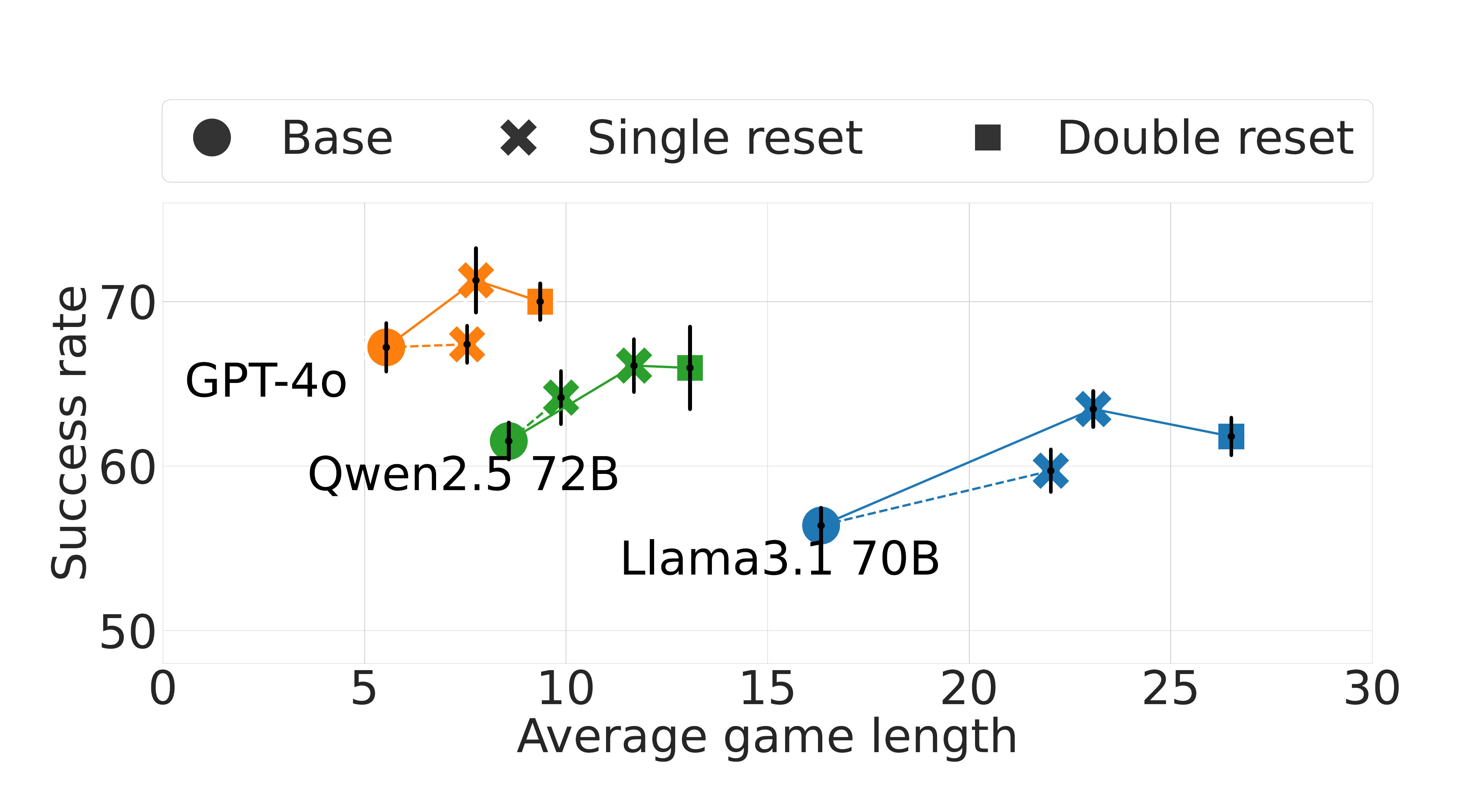}
        \caption{\textbf{Main results for \ourenvsym{}.} Dor all models, live interventions lead to significant improvement in \success{} of up to 7 points, outperforming the baseline (a cross connected by a dashed line). Interestingly, resetting twice does not offer further improvements. Black lines indicate standard error.}
        \label{fig:symmetric-env-results}
\end{figure}

\paragraph{\ourenv{}}
Fig.~\ref{fig:asymmetric-env-results} and~\ref{fig:symmetric-env-results} present the results for \ourenvasym{} and \ourenvsym{}, respectively. For \ourenvasym{} our method outperforms the base model by 6.1\%, 10.6\%, 10.3\% for \gpt{}, \llama{}, and \qwen{} respectively, and by 4.1\%, 7.0\%, 4.5\% for \ourenvsym{}. When one reset is allowed, our method outperforms the best baseline on average across models and the variants by 6.3\%.

For \ourenvasym{}, more resets yield additional gains (11.8\% vs 9.0\% with one reset across models on average). Interestingly, the open-weight \qwen{} and \llama{} perform similarly to \gpt{} with two resets.
However, a second reset did not further boost performance on \ourenvsym{}. We hypothesize that this is due to the structured form of communication in \ourenvsym{}, which would result in less communication-based mistakes.

Our interventions come at a cost of additional turns. The average number of turns increases by a factor of 1.9 and 1.6 with double and single reset for \ourenvasym{} and 1.6 and 1.4 for \ourenvsym{}.
Detailed results per variant and model can be found at \S\ref{app:fullwhodunit}.

\paragraph{\CG{}}
Tab.~\ref{tab:coding} presents the results for \llama{} on \HumanEval{} and \LCB{}, obtained with the monitors trained on \HumanEval{}. Our method reaches a Pass@1 score of 83.5\%, outperforming zero-shot prompting at 80.5\% and a multi-agent network without monitoring which reaches 81.6\%. Applying the same monitors to \LCB reaches a Pass@1 score of 21.6\%, compared to 18.2\% by zero-shot prompting and 20.4\% in the multi agent system without monitoring. This shows that not only is our method effective for realistic tasks, but also that our monitors generalize well between similar tasks.

\begin{table}[t]
\centering
\setlength\tabcolsep{3.8pt}
\footnotesize
  \begin{tabular}{lcc}
    \toprule
    \textbf{Method} & \textbf{HumanEval} & \textbf{LiveCodeBench}\\ 
    \midrule
    Zero-shot prompting & 80.5\% & 18.2\%\\ 
    Multi-agent & 81.6\% & 19.3\% \\ 
    Multi-agent + monitor & \textbf{83.5\%} & \textbf{21.8\%} \\
    \bottomrule
  \end{tabular}
  \caption{\textbf{Performance on \CG{}} measured with Pass@1. Using \llama{} with monitors trained for code generation on \HumanEval{} shows gains that generalize to a similar task in \LCB.}
  \label{tab:coding}
\end{table}

\paragraph{\govsim{}}
Tab.~\ref{tab:govsimres} presents the results for \govsim{}. With \qwenlarge{}, our method leads to an increase of 20.0\% in \survivalrate{} and 1.2 steps in \survivaltime. \efficiency{} remains similar, with 49.4\% vs 48.8\%. 
For \gpt{}, we increase \efficiency{} by 6.9\% on average, while maintaining a maximal \survivalrate{}. Overall, our method leads to better collaboration and performance on the task.

\begin{table}[t]
\centering
\setlength\tabcolsep{3.2pt}
\footnotesize
\resizebox{0.995\columnwidth}{!}{
  \begin{tabular}{lcccc}
    \toprule
    \textbf{Model} & \textbf{Ours} & \textbf{Survival} & \textbf{Survival} & \textbf{Efficiency} \\
     &  & \textbf{Rate} & \textbf{Time} & \\
    \midrule
    \textsc{Qwen} & \ding{55} & ~$35.0\pm20.1$ & ~$7.5\pm0.7$ & ~~$\mathbf{49.4}\pm 10.1$ \\
    \textsc{-1.5-110B} & \checkmark & $\mathbf{55.0}\pm21.8$ & $\mathbf{8.6}\pm0.8$ & ~$48.8 \pm 9.9$ \\
    \midrule
    \multirow{2}{*}{\gpt{}} & \ding{55} & $\mathbf{100}$ & $\mathbf{12.0}$ & ~$69.1 \pm 6.6$ \\
    & \checkmark & $\mathbf{100}$ & $\mathbf{12.0}$ & $\mathbf{76.0}\pm 4.8$ \\
    \bottomrule
  \end{tabular}
}
  \caption{\textbf{Main results for \govsim{}.} 
  Interventions lead to significant gains in \survivalrate{} and \survivaltime{} with \qwenlarge{} and in \efficiency{} with \gpt{}. Numbers include 95\% confidence intervals.}
  \label{tab:govsimres}
\end{table}

\section{Ablations and Analysis}
\label{sec:analysis}

We conduct additional analysis on \ourenvasym{}, demonstrating that our monitors capture meaningful signals of agent uncertainty and the effectiveness of our interventions.

\begin{figure}[t]
  \includegraphics[width=\columnwidth]{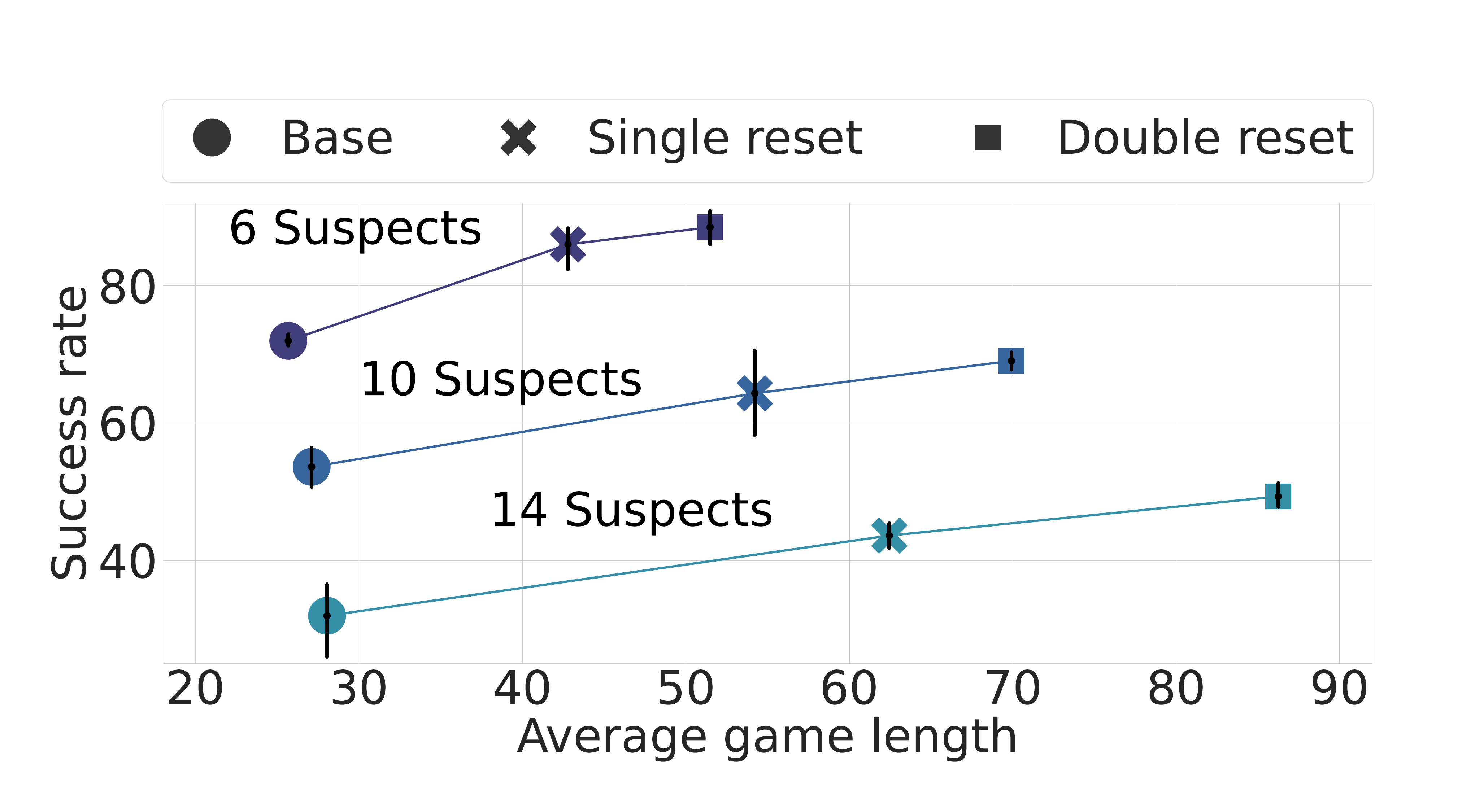}
  \caption{\textbf{Performance for \ourenvsym{} across difficulty levels}. Results are for \llama{} with 6, 10, 14 suspects. \success{} decreases with more suspects, but gains are consistent.}
  \label{fig:complexity}
\end{figure}

\begin{table}[t]
\centering
\setlength\tabcolsep{3.8pt}
\footnotesize
  \begin{tabular}{lcc}
    \toprule
    \multirow{2}{*}{\textbf{Method}} & \textbf{Success} &\textbf{Average} \\ 
    & \textbf{Rate} & \textbf{Length}\\ 
    \midrule
        Ours (Double reset) & \textbf{72.2} $\pm$ 1.6 & 42.5 $\pm$ 0.4\\ 
    Ours (Single reset) & 70.1 $\pm$ 1.7 & 31.7 $\pm$ 1.1 \\ 
    \midrule
    Monitoring: ~2nd best monitor & 69.3 $\pm$ 1.7 & 34.0 $\pm$ 1.5\\ 
    Monitoring: ~Worst monitor & 62.0 $\pm$ 1.8 & 30.6 $\pm$ 0.4 \\ 
    Intervention: Resample agent & ~~61.3 $\pm$ 1.8 & 20.1 $\pm$ 0.4\\ \midrule 
    Best baseline & 62.5 $\pm$ 1.8 & 34.3 $\pm$ 0.7\\
    No intervention & 59.8 $\pm$ 1.8 & 23.9 $\pm$ 0.5 \\

    \bottomrule
  \end{tabular}
  \caption{\textbf{Ablations for monitoring and interventions.} Strong classifiers and interventions on the communication channel are needed to improve results. Experiments are with \qwen{} and show standard error.}
  \label{tab:ablations}
\end{table}

\subsection{Ablations}

\begin{figure*}
    \centering
    \begin{minipage}[t]{0.24\textwidth}
        \centering
        \includegraphics[width=\textwidth]{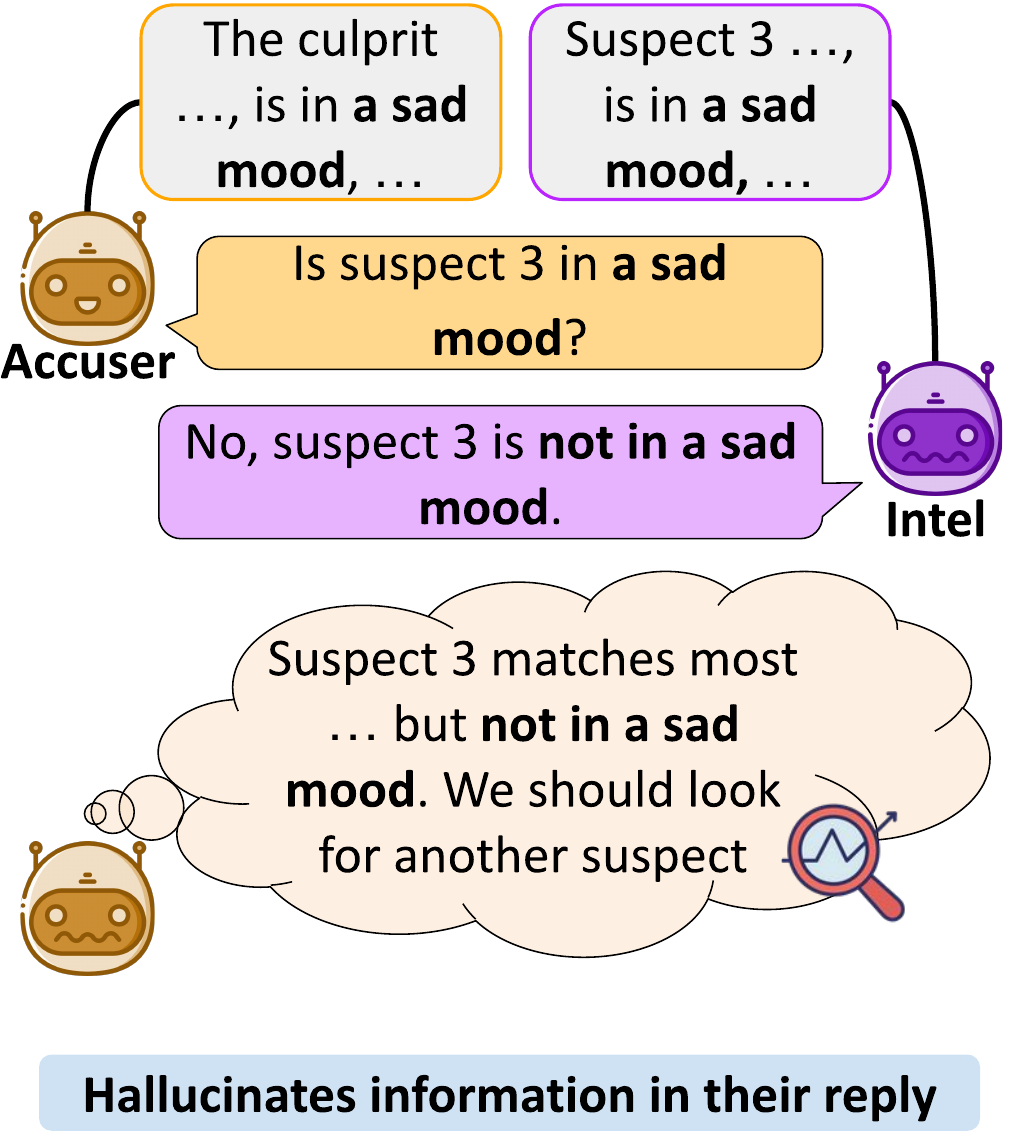}
       
        \label{fig:hallucination}
    \end{minipage}
    \hfill
    \begin{minipage}[t]{0.24\textwidth}
        \centering
        \includegraphics[width=\textwidth]{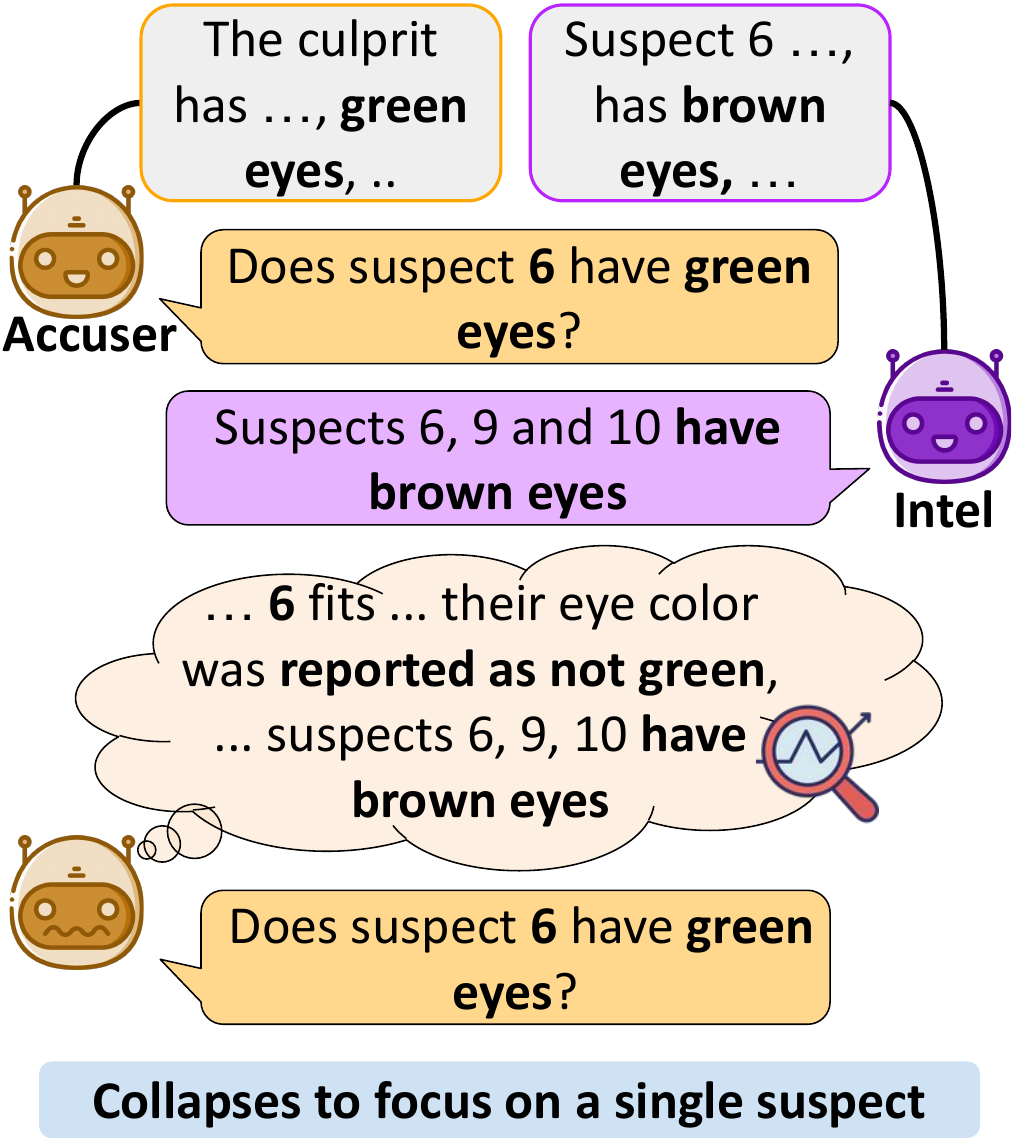}
        \label{fig:hyperfocus}
    \end{minipage}
    \hfill
    \begin{minipage}[t]{0.24\textwidth}
        \centering
        \includegraphics[width=\textwidth]{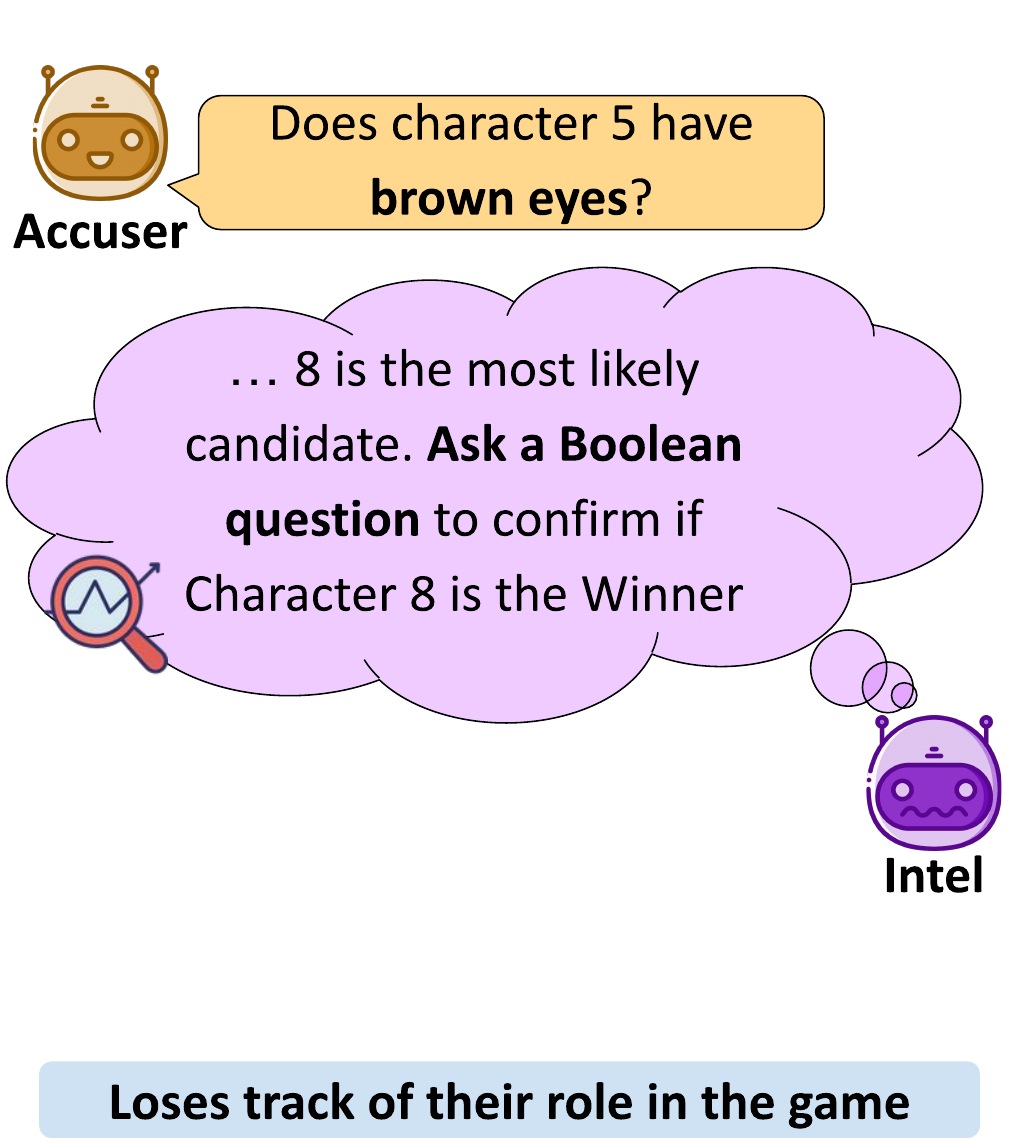}
        \label{fig:lostrole}
    \end{minipage}
    \hfill
    \begin{minipage}[t]{0.24\textwidth}
        \centering
        \includegraphics[width=\textwidth]{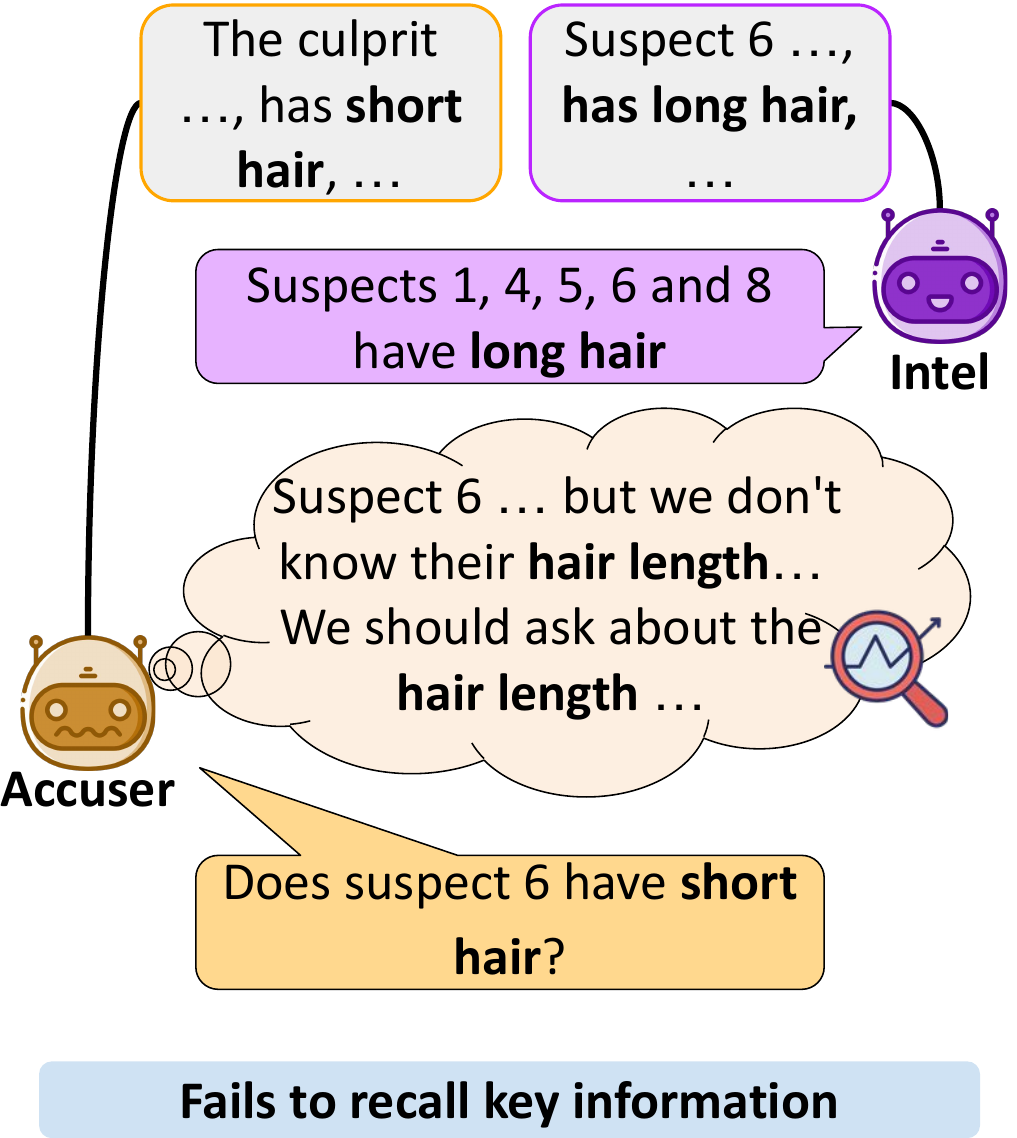}
        \label{fig:inconsistency}
    \end{minipage}
    \vspace{-15pt}
    \caption{\textbf{Example monitor triggers for \ourenvasym{} with \llama{}}. We categorize triggers into  the following categories: hallucination (48\%), agent collapse (16\%), losing track of their role (8\%) and recall failure (4\%). The information relevant to the mistake is presented in bold text.}
    \label{fig:examples}
\end{figure*}

We conduct ablations to verify improvements are a result of our monitoring and intervening approach.
For monitoring, we experiment with the classifiers that performed second best or the worst on the validation set to verify that classifier performance is correlated with final gains.
For interventions, we experiment with resampling the agent upon trigger instead of resetting the communication channel.

Tab.~\ref{tab:ablations} presents the results for \qwen{}.
For monitoring, the second best monitor leads to slightly lower gains (9.5\% vs 10.3\%) compared to the worst monitor with only 2\% improvement, which is similar to the baseline at 2.6\%. 
For interventions, resampling a single agent is ineffective, with a gain of 1.5\%, which is 1.5\% below the best random baseline. 
Overall, these results suggest that strong monitors and interventions are needed to improve performance.

\subsection{Performance by Task Complexity}
Fig.~\ref{fig:complexity} presents the results for \llama{} on \ourenvasym{} with 6, 10, and 14 suspects, using the monitor trained for 10 suspects.
We observe consistent gains of 14.0\%, 10.7\% and 11.7\% in \success{} with a single reset and 16.5\%, 15.4\% and 17.4\% with two resets across the different complexities. The average game length increases with complexity, from 1.9 with six suspects to 3.1 with fourteen suspects.
This shows our method generalizes across task complexities with no required monitor retraining. Similar analysis for \ourenvsym{} is provided in \S\ref{app:complex_symm}.

\subsection{Qualitative Analysis}

To understand what phenomena are captured by our monitors, we perform a qualitative analysis of 50 examples in which monitors were triggered in \ourenvasym{} with \llama{}. 
We observed four categories of monitor triggers: (a) \emph{hallucination} (48\% of reviewed cases): the information shared was incorrect, (b) \emph{collapse} (16\%): an agent repeatedly asks about the same suspect, even when they already received the relevant information, (c) \emph{role loss} (8\%): the agent loses track of their role in the game, and (d) \emph{recall failure} (4\%): Accuser failed to recall key information that was previously shared by the Intel.
Fig.~\ref{fig:examples} presents an example for each category.
In the remaining cases, we could not identify a concrete issue in the game trajectory, suggesting they may be instances of a \emph{false trigger}.

Overall, 76\% of the monitor triggers belong to one of the four main categories, with hallucinations being the most frequent trigger.
Our analysis shows that our monitors successfully detect a wide range of problematic trajectories, and gains are due to high monitoring accuracy in intervened games.

%% file: latex/sections/06_related_work.tex
\section{Related Work}
\label{sec:rw}

\paragraph{Multi-agent communication} 
Communication between agents have long been suggested in order to develop strong modular systems \cite{KRAUS199779, 8352646, sukhbaatar2016learningmultiagentcommunicationbackpropagation, foerster2016learningcommunicatedeepmultiagent, 10.1007/978-3-319-75477-2_2, lazaridou2020emergentmultiagentcommunicationdeep, lowe2020multiagentactorcriticmixedcooperativecompetitive}. 
One of the most exciting applications of language agents is environments where agents autonomously communicate, with examples of improving factuality and reasoning via agent debate \cite{du2023improvingfactualityreasoninglanguage, liang-etal-2024-encouraging}, cooperation between embodied robots \cite{mandi2023rocodialecticmultirobotcollaboration, chen2024scalablemultirobotcollaborationlarge}, simulating software development teams \cite{li2023camel, hong2024metagpt, qian-etal-2024-chatdev, liu2024a}, and gaming environments \cite{mukobi2023welfarediplomacybenchmarkinglanguage, xu2024exploringlargelanguagemodels}.
Thus, several frameworks have been proposed to enable simple development of multi-agent environments \cite{li2023camel, wu2023autogenenablingnextgenllm, hong2024metagpt}. 
We contribute to this line of work by introducing \ourenv{}, a modular environment for multi-agent communication, which allows easy configuration of difficulty levels and communication structures.

\paragraph{Robustness of multi-agent systems} Previous work has showed that multi-agent collaboration is susceptible to adversarial attacks \cite{wang2025gsafeguardtopologyguidedsecuritylens, amayuelas-etal-2024-multiagent} and that Theory of Mind can improve collaboration in simple gaming environments \cite{lim2020improvingmultiagentcooperationusing}. Our method of monitoring and intervention 
can complement existing methods by improving agent communication,
and potentially prevent attacks such as poisoned information or prompt injections \cite{nazary2025poisonragadversarialdatapoisoning, greshake2023youvesignedforcompromising}, which can be viewed as agents momentarily going \emph{rogue}.

\paragraph{Uncertainty estimation in language modeling} Uncertainty estimation has been shown useful in detecting and mitigating hallucinations in knowledge-intensive tasks \cite{kadavath2022languagemodelsmostlyknow, yona-etal-2024-large, ivgi2024from}, including in \emph{agentic retrieval}  where an external search engine is used \cite{jiang-etal-2023-active, han-etal-2024-towards}.
It has also been recently applied to language agents in order to increase exploration \cite{rahn2024controlling}, improve performance on bandit tasks \cite{felicioni2024on}, or make agents output textual estimates to help debates \cite{debunc}. In this work, we bridge uncertainty estimation and multi-agent collaboration by training monitors to predict the probability of task failures given agents uncertainty.

\paragraph{Aggregating multiple generations}

Aggregating over multiple generations is a popular method to increase performance \cite{wang2023selfconsistency, yoran-etal-2023-answering, du2023improvingfactualityreasoninglanguage, chen2024universal, min2024beyond}.
However, post-hoc aggregation is not directly applicable in agentic settings, where actions can be \emph{irreversible}.
Additionally, majority voting \cite{wang2023selfconsistency} requires at least three generations, while our approach increases turn account by less than twofold on average in our main experiments with \ourenv{}.
Our work differs by resetting the communication channel \emph{before} a problematic action was taken, rather than aggregating \emph{after} the final prediction.

%% file: latex/sections/07_conclusion.tex
\section{Conclusion}
We present live monitoring and mid-run interventions for multi-agent systems. We demonstrate that monitors based on simple statistical measures can effectively predict future agent failures, and these failures can be prevented by restarting the communication channel. Experiments across multiple environments and models show consistent gains of up to 17.4\%-20\% in system performance, with an addition in inference-time compute.
Our work also introduces \ourenv{}, a new environment for studying multi-agent cooperation.

%% file: latex/sections/limitations.tex
\section*{Limitations}

\paragraph{Environments} In this work, we focus on \ourenv{}, a new framework for multi-agent communications, in addition to GovSim \cite{piatti2024cooperate}. We believe that adapting our approach to additional environments is an exciting direction for future work.
We note that our approach can be directly extended only to environments where agents achieve non-zero accuracy, as otherwise interventions will have no affect on performance.
In addition, we showed that intervening by restarting an agent without the communication is less effective (\S\ref{sec:analysis}).
Our work can potentially be extended to single-agent settings by developing novel monitors and interventions.

\paragraph{Sample size}
For \ourenv{}, we experiment with 180 test environments . While this is a relatively small number, it is in line with recent evaluation sets that have 125-300 examples such as HumanEval \cite{chen2021evaluatinglargelanguagemodels}, DrawBench \cite{saharia2022photorealistic}, Bamboogle \cite{press-etal-2023-measuring}, and SWE-Bench \cite{jimenez2024swebench}. For \govsim{}, we extend from a single starting resource $R_0=100$ in \cite{piatti2024cooperate} to 20 choices. 
To reduce noise, we run every experiment between three and four times and report confidence intervals or standard deviation.
Nevertheless, smaller evaluation sets also have some advantage as they require less compute to evaluate, thus having a smaller environment footprint \cite{10.1145/3381831} and allowing more research teams to experiment with our environments.  

\paragraph{Data collection for training monitors}
To train our monitors to predict if an intermediate state of an agent will result in a task failure, we assume access to a set of train tasks. Realistically, collecting labels can be expensive, especially when task failures incur high costs. An exciting direction for future work is to frame our monitors as an online learning problem \cite{LITTLESTONE1994212, 10.5555/3041838.3041955, park2024llmagentsregretcase}, where one has to learn to detect rogue agents whilst minimize the number of system failures.

\section*{Ethical Implications and Broader Impact}
While multi-agent collaboration is an exciting direction for future research, it also entails significant risks.
Strong multi-agent systems can potentially solve tasks beyond the reach of current AIs, a potential risk if used by a malicious user.
In addition, multi-agent collaboration can have major economic and social impact. For example, a strong multi-agent system for autonomous software development can significantly increase productivity of engineering teams.

Another exciting future direction for multi-agent collaborations is operating embodied robots \cite{mandi2023rocodialecticmultirobotcollaboration, chen2024scalablemultirobotcollaborationlarge} and simulating human behavior \cite{10.1145/3586183.3606763}, which can cause significant social impact. In these domains, monitoring when future errors are likely and intervening when monitors are triggered can be especially important, as safe deployment presents a major challenge.
Moreover, environmental considerations should also be taken into account, as multi-agent systems often incur high costs (seen from our experiments in \ref{app:models}).

%% file: latex/sections/appendix.tex
\section{Uncertainty estimation for monitors}
\label{sec:stat_measures}
In our monitoring, we use 3 statistical measures over the probability vector: Entropy, Varentropy and Kurtosis.
Given a probability vector $P$, Entropy, Varentropy and Kurtosis are defined as:
\begin{align*}
&\text{Entropy}(P) = -\sum_{p\in P} log(p)\cdot p \\
&\text{Varentropy}(P) = \\ 
&\hspace{1cm} -\sum_{p\in P} p \cdot (log(p) + \text{Entropy}(P))^2 \\
& \text{Kurtosis}(P) = \\
&\hspace{1cm} \frac{\sum_{p\in P} p (-log(p) - \text{Entropy}(P))^4}{\text{Varentropy}(P)^2}
\end{align*}

\section{\ourenv{}}
\label{app:attributes}
\paragraph{Attribute list} The suspects in \ourenv{} are defined by a collection of different attributes. These attributes are modular and can be swapped as need arises. In our experiments, we used 12 attributes for the asymmetric variant and 20 attributes for the symmetric variant. This was done to create a challenging yet feasible task for \ourenvsym{}, which is easier. The complete list of attributes and their possible values is given in Tab.~\ref{tab:attributes}:

\begin{table}[t]
  \centering
  \footnotesize 
  \begin{tabular}{lccc}
    \toprule
    \textbf{Attribute} & \textbf{Values} \\
    \midrule
    hat & brown, black \\
    mood & happy, sad \\
    shirt color & pink, green \\
    hobby & basketball, dancing \\
    pants & long, short \\
    pants color & brown, black \\
    eye color & blue, brown, green \\
    eye glasses & circular, square \\
    shirt & button-up, tee \\
    shoe color & red, white \\
    hair & long, short \\
    watch & bronze, silver \\
    \midrule
    socks & dotted, white \\
    jacket & yellow, jean \\
    height & short, tall \\
    age & young, old \\
    build & medium, muscular \\
    personality & introverted, extroverted \\
    interests & sports, arts \\
    occupation & professional, student \\
    \bottomrule
  \end{tabular}

  \caption{\textbf{List of attributes used in the \ourenv{} variants experiments}. The top half appears in experiments of both variants, while the bottom one only appears in the symmetric variant experiments. This list is completely modular and can be changed to affect task complexity.}
  \label{tab:attributes}
\end{table}

\section{\CG{}}
\label{app:codegen}
\CG{} is a multi-agent coding environment, heavily drawing from the debate frameworks presented in \citet{du2023improvingfactualityreasoninglanguage} and \citet{liu2024dynamicllmpoweredagentnetwork}. Parts of the environment code of \citet{liu2024dynamicllmpoweredagentnetwork} was used to construct our coding generation debate environment, \CG{}, including the coding role implementation of `PythonAssistant', `AlgorithmDeveloper', `ComputerScientist' and `Programmer'. The dynamic network aspect of \citet{liu2024dynamicllmpoweredagentnetwork} was not used in our experiments. 

\section{Experiments}
\subsection{Data Splits}
\label{app:datasplits}
For \ourenv{} we define games by suspect attributes. For \govsim{} we make use of the starting resource, $R_0$. For Train and validation we shuffle and split $\{105 + 5k | k\in [0, 19]\}$ into two exclusive parts. Test is simply $R_0\in \{[100], [210,300]\}$. We make sure $R_0=100$ stays in the test set to align with results from \citet{piatti2024cooperate}.

\subsection{Monitor Design}
\label{app:classifiers}
The monitor is made up of a classifier function together with a threshold.
To train the classifier we first split the data set into a train set and a validation set. We use the training set to min-max normalize the data to the [-1, 1] range, then create polynomial features over input features with degree D, and finally fit a Ridge function over the resulting polynomials. These are done using the scikit-learn library  \cite{scikit-learn}. We do this for $D\in[1,2,3,4,5]$ and for every combination of the 3 features. Turn count is always included in the model input.
To find the optimal degree, feature combination and final classifier threshold $\tau \in [0,1]$, we use the validation set. 

For every game, we use the monitor to test triggers against the golden task success labels from the run. We consider the monitor successful on a game if even one sample of a failed game is over the threshold, since that would trigger a reset. We consider the monitor unsuccessful on a game if either the classifier was never over the threshold for a failed game, or was over the threshold for any turn on a successful game. Finally, we pick the monitor which had the maximum \textbf{gain} over the validation set. This would be the monitor we use for testing. See Tab.~\ref{tab:classifiers_used} for monitors used in \ourenvasym{} and their predicted gain. \S\ref{app:triggers} shows a histogram of monitor turn triggers in a game without interventions.

For baselines, we used a random monitor with a set probability of triggering $p$. We applied the monitor to the validation set and chose $p\in[0,1]$ that maximized gains.

\begin{table}[t]
\centering
\setlength\tabcolsep{3.8pt}
\footnotesize
  \begin{tabular}{lccccc}
    \toprule
    \textbf{Model} & \textbf{Type} & \textbf{Features} & \textbf{Degree} & \textbf{Gain} & $\tau$\\
    \midrule
    \textsc{Qwen-2.5} & Accuser & Var & 1 & 5.8 & 0.51\\
    \hspace{0.5cm}\textsc{-72B} & Intel & Ent & 1 & 4.6 & 0.36 \\
    \midrule
    \textsc{Llama-3.1} & Accuser & Var Kur & 5 & 11.6 & 0.55\\
    \hspace{0.5cm}\textsc{-70B} & Intel & Ent Var & 4 & 11.3 & 0.5 \\  
    \midrule
    \multirow{2}{*}{\textsc{\gpt{}}} & Accuser & Ent & 3 & 4.9 & 0.37 \\
    & Intel & Ent Var & 5 & 5.3 & 0.43\\
    \bottomrule
  \end{tabular}
  \caption{\textbf{\ourenvasym{} best monitors}. Best monitors for different models had different features and degrees selected.}
  \label{tab:classifiers_used}
\end{table}

\begin{table}[t]
\centering
\setlength\tabcolsep{4pt}
\footnotesize
  \begin{tabular}{lp{4.5cm}}
    \toprule
    \textbf{Model} & \textbf{Full name} \\
    \midrule
    \textsc{\qwen{}} & Qwen/Qwen2.5-72B-Instruct \\
    \textsc{\llama{}} & meta-llama/Llama-3.1-70B-Instruct \\
    \textsc{\gpt{}} & gpt-4o-2024-08-06 \\
    \textsc{\qwenlarge{}} & Qwen/Qwen1.5-110B-Chat-GPTQ-Int4 \\
    \bottomrule
  \end{tabular}
  \caption{\textbf{Full model names used across experiments.}}
  \label{tab:models_used}
\end{table}

\subsection{Models}
\label{app:models}
For the open-source models, we required at most 320GB of GPU memory to load the model and run the experiments. A load of 180 games took on average 6 hours. Using \gpt{} took about one hour for 180 games, and cost about 30\$.
When using \govsim{}, \qwenlarge{} required 160GB of total GPU memory, and each game took about 3 hours of run time (depending on how long the agents last in the environment). \gpt{} took between one and two hours per game, and cost about ~8\$ per a game. See table \ref{tab:models_used} for specific model names (HuggingFace for open weights). We make use of the vLLM library for open source model loading \cite{kwon2023efficient}. All model queries are done with the default model parameters, with the exception of temperature which is always set to 0.

\subsection{Full Experimental Results}
\label{app:fullwhodunit}
Due to the high variance we encountered during runs, we run a total of 180 test games for each variant of \ourenv{}. Due to cost, we only run \gpt{} three times on each game. The full results of each experiment, including \success{} and \precision{} can be seen at Tab.~\ref{tab:asymmresults} for the asymmetric variant and at Tab.~\ref{tab:symmresults} for the symmetric.

\subsection{\ourenvsym{} complexity analysis}
\label{app:complex_symm}
\begin{figure}[t]
  \includegraphics[width=\columnwidth]{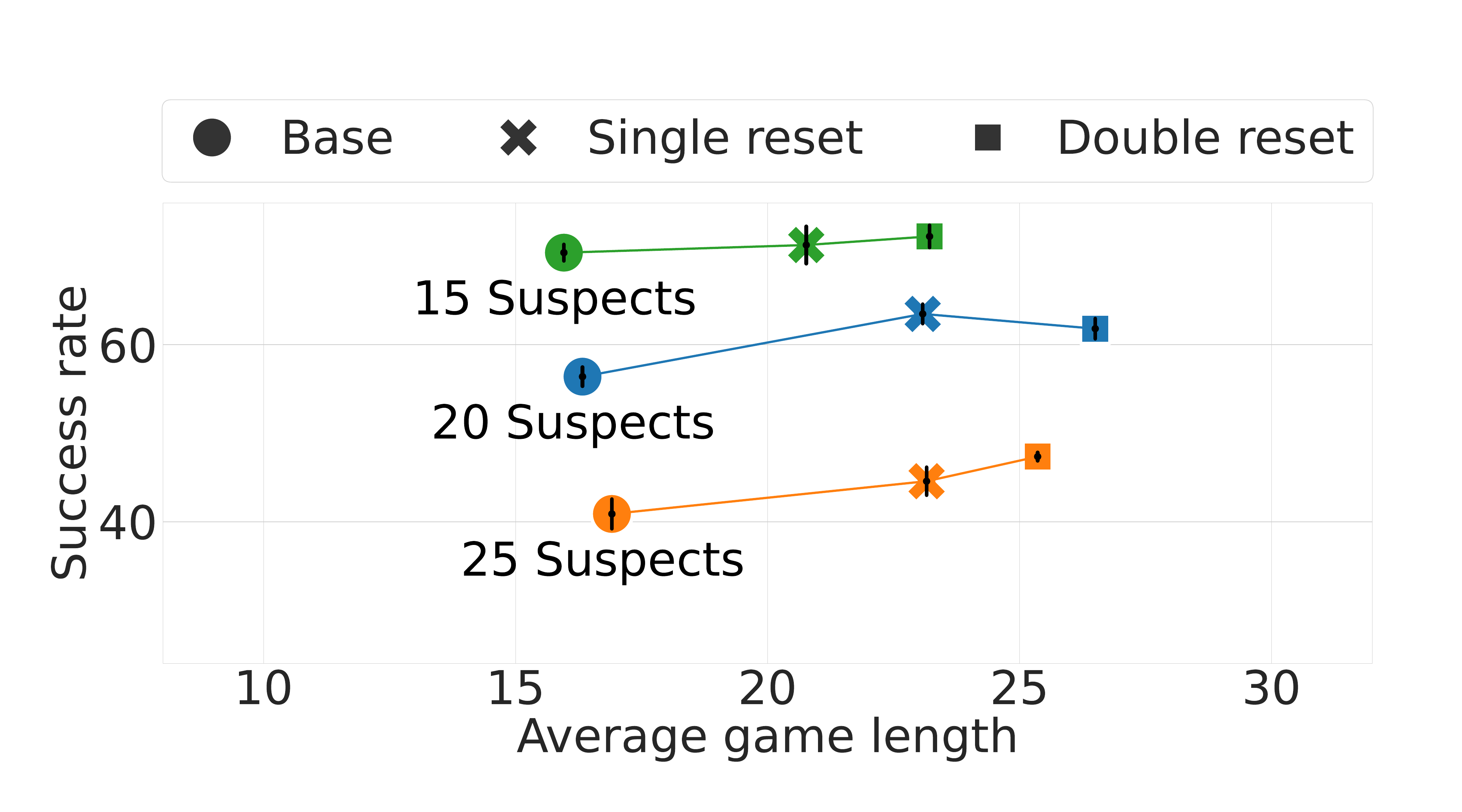}
  \caption{\textbf{Performance for \ourenvsym{} across difficulty levels}. Results are for \llama{} with 15, 20, 25 suspects. \success{} decreases with more suspects. Gains with 15 suspects is negligible, however gains with 25 suspects are similar to the original 20 suspect setting.}
  \label{fig:complexity_symm}
\end{figure}

As a complement to the \ourenvasym{} complexity analysis seen in \S\ref{sec:results}, Fig.~\ref{fig:complexity_symm} shows a complexity analysis of \ourenvsym{}.
Since our main results for \ourenvsym{} have 20 suspects (see \S\ref{sec:experimental_setting}), we conduct analysis with an easier 15 suspect setting and a harder 25 suspect setting. Similarly to \ourenvasym{}, we use the original monitor, trained with 20 suspect games. 
After a single reset, the 15 and 25 suspect settings saw a gain of 0.86\% and 3.69\% over their base experiments, respectively. With an additional reset, the 15 suspect setting only gained 1.83\% over the base, while the 25 suspect setting saw an increased gain of 6.47\%. The 25 suspect setting compares to the 20 suspect setting, which has an increase of 7.08\% and 5.42\%. Interestingly, the decrease seen after a double reset was not seen in the 25 suspect setting. The negligible gains in the 15 suspect setting fit with our analysis that the symmetric environment is easier when the number of suspects is small, hinting that communication problems are not very common in this setting.

\begin{table}
\centering
  \small
  \begin{tabular}{|c | c c c|}
    \toprule
    \multirow{2}{*}{\textbf{Model}} & \multirow{2}{*}{\textbf{Int. \#}} & \textbf{Success{}} & \multirow{2}{*}{\textbf{\precision{}}} \\
     &  & \textbf{-Rate{}} &  \\
    
\midrule
\multirow{16}{*}{\qwen{}} & \multirow{4}{*}{0} & 57.8 & 73.8 \\
 & & 62.8 & 79.0 \\
 & & 58.9 & 81.5 \\
 & & 60.0 & 77.1 \\
\cline{2-4}
 & \multirow{4}{*}{1} & 67.8 & 78.2 \\
 & & 73.9 & 87.5 \\
 & & 66.7 & 79.5 \\
 & & 72.2 & 81.2 \\
\cline{2-4}
 & \multirow{4}{*}{2}  & 67.2 & 72.9 \\
 & & 72.8 & 77.5 \\
 & & 76.7 & 80.7 \\
 & & 72.2 & 77.4 \\
\cline{2-4}
 & \multirow{4}{*}{Baseline} & 58.9 & 72.6 \\
 & & 63.9 & 76.7 \\
 & & 63.9 & 81.6 \\
 & & 63.3 & 78.1 \\
\midrule
\multirow{16}{*}{\llama{}} & \multirow{4}{*}{0} & 56.1 & 73.7 \\
 & & 51.1 & 69.2 \\
 & & 50.6 & 78.4 \\
 & & 56.7 & 78.5 \\
\cline{2-4}
 & \multirow{4}{*}{1} & 55.6 & 71.4 \\
 & & 66.1 & 75.8 \\
 & & 62.2 & 74.2 \\
 & & 73.3 & 82.0 \\
\cline{2-4}
 & \multirow{4}{*}{2} & 69.4 & 77.2 \\
 & & 67.8 & 75.8 \\
 & & 71.1 & 77.6 \\
 & & 67.8 & 77.2 \\
\cline{2-4}
 & \multirow{4}{*}{Baseline} & 53.9 & 75.2 \\
 & & 57.8 & 72.2 \\
 & & 54.4 & 74.2 \\
 & & 52.8 & 72.0 \\
\midrule
\multirow{12}{*}{\gpt{}} & \multirow{3}{*}{0} & 63.3 & 63.7 \\
 & & 61.7 & 61.7 \\
 & & 61.1 & 61.5 \\
\cline{2-4}
 & \multirow{3}{*}{1} & 70.0 & 70.4 \\
 & & 67.2 & 68.4 \\
 & & 67.8 & 67.8 \\
\cline{2-4}
 & \multirow{3}{*}{2} & 70.9 & 66.1 \\
 & & 69.8 & 65.0 \\
 & & 67.0 & 62.6 \\
\cline{2-4}
 & \multirow{3}{*}{Baseline} & 61.7 & 61.7 \\
 & & 63.3 & 63.3 \\
 & & 62.8 & 62.8 \\
    
    \bottomrule
  \end{tabular}

  \caption{\textbf{List of different experiments done with \ourenvasym{}} comparing normal running, using our method for one and two resets, and a baseline random classifier that can reset once. Each of these is an average over the exact same 180 games, which were defined with 10 suspects. For each run we bring the \success{} and \precision{}.}
  \label{tab:asymmresults}
\end{table}

\begin{table}
\centering
  \small
  \begin{tabular}{|c| c c c|}
    \toprule
    \multirow{2}{*}{\textbf{Model}} & \multirow{2}{*}{\textbf{Int. \#}} & \textbf{Success{}} & \multirow{2}{*}{\textbf{\precision{}}} \\
     &  & \textbf{-Rate{}} &  \\
    
    \midrule
    \multirow{16}{*}{\qwen{}} & \multirow{4}{*}{0} & 63.3 & 64.0 \\
     & & 58.3 & 58.3 \\
     & & 62.8 & 63.5 \\
     & & 61.7 & 62.0 \\
    \cline{2-4}
    
     & \multirow{4}{*}{1} & 62.2 & 62.9 \\
     & & 70.0 & 70.4 \\
     & & 65.6 & 65.9 \\
     & & 66.7 & 67.0 \\
    \cline{2-4}
    
     & \multirow{4}{*}{2} & 71.1 & 71.9 \\
     & & 69.4 & 69.4 \\
     & & 61.7 & 62.4 \\
     & & 61.7 & 61.7 \\
    \cline{2-4}
    
     & \multirow{4}{*}{Baseline} & 67.2 & 67.2 \\
     & & 66.7 & 67.0 \\
     & & 61.7 & 62.0 \\
     & & 61.1 & 61.5 \\
    
    \midrule
    \multirow{16}{*}{\llama{}} & \multirow{4}{*}{0} & 53.9 & 57.7 \\
     & & 55.6 & 59.9 \\
     & & 57.2 & 61.3 \\
     & & 58.9 & 63.5 \\
    \cline{2-4}
    
     & \multirow{4}{*}{1} & 61.7 & 64.5 \\
     & & 66.7 & 71.9 \\
     & & 62.8 & 64.9 \\
     & & 62.8 & 64.9 \\
    \cline{2-4}
    
     & \multirow{4}{*}{2} & 63.9 & 65.0 \\
     & & 63.3 & 65.5 \\
     & & 58.9 & 59.9 \\
     & & 61.1 & 63.6 \\
    \cline{2-4}
    
     & \multirow{4}{*}{Baseline} & 60.0 & 65.1 \\
     & & 62.2 & 66.7 \\
     & & 56.1 & 61.6 \\
     & & 60.6 & 62.3 \\
    
    \midrule
    
    \multirow{12}{*}{\gpt{}} & \multirow{3}{*}{0} & 65.0 & 65.0 \\
     & & 70.0 & 70.0 \\
     & & 66.7 & 66.7 \\
    \cline{2-4}
    
     & \multirow{3}{*}{1} & 68.3 & 68.3 \\
     & & 75.0 & 75.0 \\
     & & 70.6 & 70.6 \\
    \cline{2-4}
    
     & \multirow{3}{*}{2} & 68.9 & 68.9 \\
     & & 72.2 & 72.2 \\
     & & 68.9 & 68.9 \\
    \cline{2-4}
     & \multirow{3}{*}{Baseline} & 69.4 & 69.4 \\
     & & 67.2 & 67.2 \\
     & & 65.6 & 65.6 \\
    \bottomrule
  \end{tabular}

  \caption{\textbf{List of different experiments done with \ourenvsym{}}  using our method, including no resets and baseline random classifiers. Each of these is an average over the exact same 180 games, which were defined with 20 suspects. For each run we bring the \success{} and \precision{}. }
  \label{tab:symmresults}
\end{table}

\section{Analysis}
\subsection{Monitor trigger turns}
\label{app:triggers}
\begin{figure}[t]
\centering
  \includegraphics[width=\columnwidth]{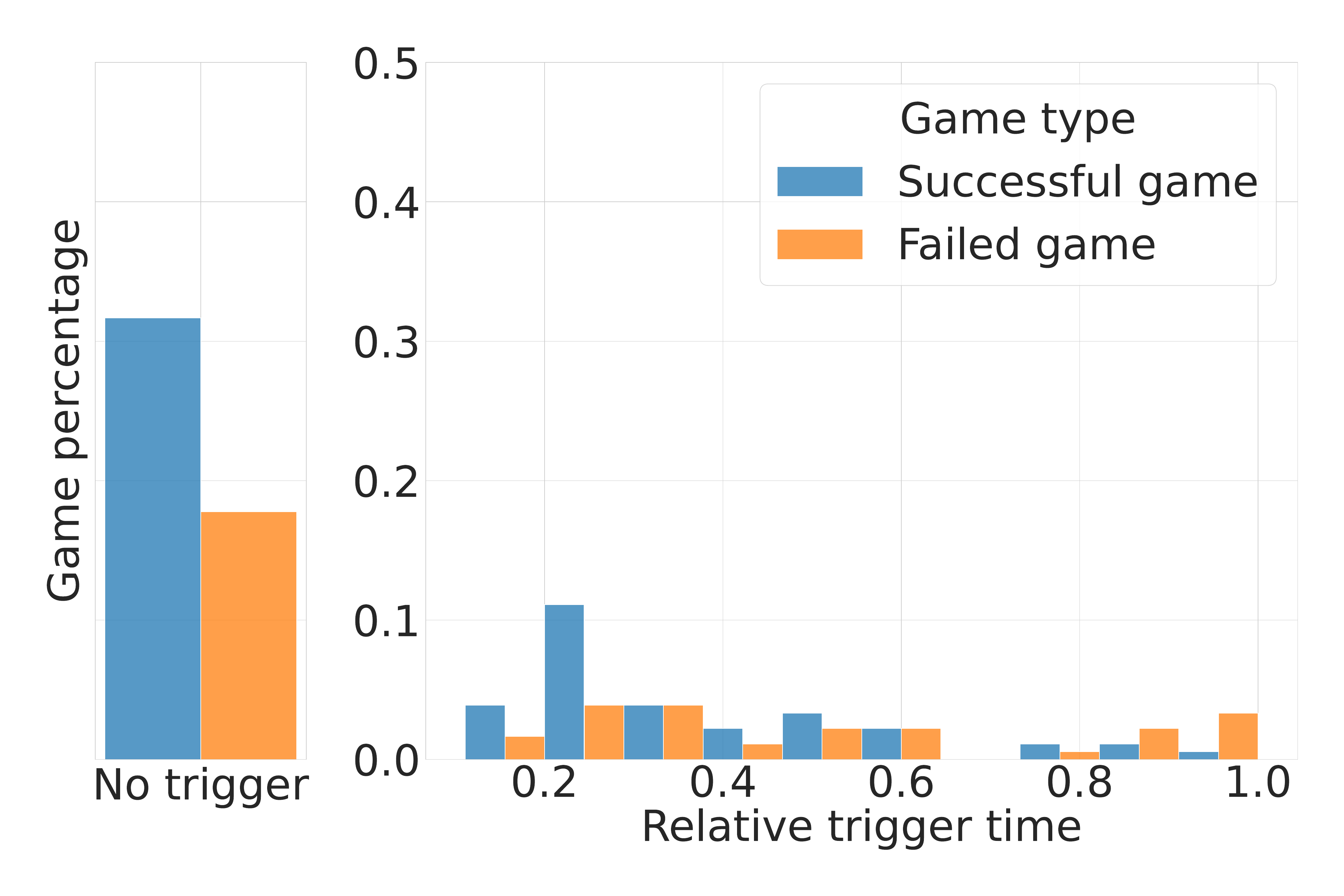}
  \caption{\textbf{Histogram of monitor triggers} for the \ourenvasym{} test set using \gpt{}, normalized by the 180 game count.}
  \label{fig:triggers}
\end{figure}

Fig.~\ref{fig:triggers} shows a histogram of the turn count for a monitor trigger on \ourenvasym{}, across all 3 models and all experiments. We see that triggers when the game ended up succeeding usually happen early. We hypothesize these are a result of a communication mistake that agents manged to recover from. On the other hand, communication mistakes later in the conversation often lead to task failure. 20\% of games failed without a monitor trigger, leaving us further room for monitor improvement and gain.

\section{\ourenv{} Prompts}
\label{app:prompts}
We make here available the full text of \ourenv{} prompts.
For \ourenvsym{} this includes System prompts for all agents (Fig.~\ref{fig:sym_sysprompts}), user prompts (Fig.~\ref{fig:sym_userprompts}), and message formats (Fig.~\ref{fig:sym_formats}).
For \ourenvasym{} this includes System prompts for Accuser (Fig.~\ref{fig:asym_sysprompts}) and Intel (Fig.~\ref{fig:asym_sysprompts_intel}), User prompts (Fig.~\ref{fig:asym_userprompts}) and message formats (Fig.~\ref{fig:asym_formats}).

\begin{figure*}[!ht]
    \centering
    \begin{tcolorbox}[colback=gray!10,arc=0pt,outer arc=0pt,boxrule=0pt,toprule=0.5mm,bottomrule=0.5mm,rightrule=0.5mm,leftrule=0.5mm] 
        \small
        \begin{minipage}{\linewidth} 
        
            \begin{tcolorbox}[colback=blue!5,
            colframe=blue!50,arc=0pt,outer arc=0pt,boxrule=0pt,toprule=0.5mm,bottomrule=0.5mm,rightrule=0.5mm,leftrule=0.5mm, title=System Prompt,coltitle=black]
                \textbf{Introduction:} \\
                - Your name is NAME, and you are a helpful AI assistant participating in a cooperative game with other players. There are \texttt{AGENT\_COUNT} players in total: \texttt{AGENT\_NAMES}. \\

                \textbf{Game Overview:} \\
                - The game involves characters, one of which is the Winner and will receive an award! Each player has information about all characters and partial knowledge about who the Winner is. \\
                - You will receive information in the following format: \\
                \begin{enumerate}
                    \item A list of characters with their descriptions.
                    \item A list of starting facts about the Winner, describing specific properties. These are only visible to you.
                    \item The current communication channel between you and other players, containing shared facts.
                \end{enumerate}

                \textbf{Game Mode:} \\
                - The game proceeds in rounds. In each round, you can either award the prize to a character you believe is the Winner, share one of your starting facts or skip your turn. \\
                - If there is only one character matching the criteria according to your information, you should award them the prize. Otherwise, if there's a fact that adds to your team's understanding, you should share a fact. If you believe you have nothing to add, you can skip your turn. \\
                - The goal is to find the correct Winner in the minimum number of rounds. Awarding an incorrect Winner will lose points. \\
                - \textbf{Output Format in Game Mode:} \\
                \begin{verbatim}
{
  "thoughts": "Thoughts describing your optimal action at this point. 
  Describe whether to share a fact, award a character or skip your turn,
  and why you chose this action.",
  "action": 1, 2 or 3,  // 1 to share a fact, 2 to award a character 
  or 3 to skip your turn
  "fact": If sharing, fact number to share. This must be a number.
  "character": If accusing, character number of accuse. This must be a number.
}
                \end{verbatim}

                \textbf{Instructions Recap:} \\
                - Focus on minimizing rounds to find the Winner. \\
                - Share informative facts to narrow down the possibilities. \\
                - Award a character only when sure of their identity as the Winner. \\
                - Follow the specified JSON format for output in Game mode. \\

                \texttt{SUSPECT\_INFORMATION}
            \end{tcolorbox} 
        \end{minipage}
    \end{tcolorbox}
    \caption{\textbf{
    \ourenvsym{} system prompts.}}
    \label{fig:sym_sysprompts}
\end{figure*}

\begin{figure*}[!ht] 
    \centering
    \begin{tcolorbox}[colback=gray!10,arc=0pt,outer arc=0pt,boxrule=0pt,toprule=0.5mm,bottomrule=0.5mm,rightrule=0.5mm,leftrule=0.5mm] 
        \small
        \begin{minipage}{\linewidth} 
            \begin{tcolorbox}[colback=blue!5,
            colframe=blue!50, arc=0pt,outer arc=0pt,boxrule=0pt,toprule=0.5mm,bottomrule=0.5mm,rightrule=0.5mm,leftrule=0.5mm, title=User prompt,coltitle=black]
                You received the following starting facts about the Winner (these are the facts you may share): \\
                \texttt{FACTS} \\

                The current communication channel between you and the other players is: \\
                \texttt{COMM\_CHANNEL} \\

                The current turn is \texttt{TURN\_COUNT} out of \texttt{MAX\_TURN\_COUNT}. \\
                It’s your turn. What would you do?
            \end{tcolorbox} 
        \end{minipage}
    \end{tcolorbox}
    \caption{\textbf{
    \ourenvsym{} user prompts.}}
    \label{fig:sym_userprompts}
    
\end{figure*}

\begin{figure*}[!ht] 
    \centering
    \begin{minipage}{\linewidth} 
        \begin{tcolorbox}[colback=gray!10,arc=0pt,outer arc=0pt,boxrule=0pt,toprule=0.5mm,bottomrule=0.5mm,rightrule=0.5mm,leftrule=0.5mm] 
            \begin{minipage}{\linewidth}
                \begin{tcolorbox}[colback=blue!5,
            colframe=blue!50,arc=0pt,outer arc=0pt,boxrule=0pt,toprule=0.5mm,bottomrule=0.5mm,rightrule=0.5mm,leftrule=0.5mm, title=User Prompt and Messages,coltitle=black]

                     \begin{tcolorbox}[colback=white!5,
            arc=0pt,outer arc=0pt,boxrule=0pt,toprule=0.2mm,bottomrule=0.2mm,rightrule=0.2mm,leftrule=0.2mm]
                You received the following starting facts about the Winner (these are the facts you may share): \\
                \texttt{FACTS} \\

                The current communication channel between you and the other players is: \\
                \texttt{COMM\_CHANNEL} \\

                The current turn is \texttt{TURN\_COUNT} out of \texttt{MAX\_TURN\_COUNT}. \\
                It’s your turn. What would you do?
            \end{tcolorbox}
            
                    \begin{tcolorbox}[colback=white,arc=0pt,outer arc=0pt,boxrule=0.2mm,toprule=0.2mm,bottomrule=0.2mm,rightrule=0.2mm,leftrule=0.2mm]
                        Player \texttt{NAME} has decided to share a fact about the Winner: \texttt{FACT}.
                    \end{tcolorbox}

                    \begin{tcolorbox}[colback=white,arc=0pt,outer arc=0pt,boxrule=0.2mm,toprule=0.2mm,bottomrule=0.2mm,rightrule=0.2mm,leftrule=0.2mm]
                        The winner is \texttt{VALUE}.
                    \end{tcolorbox}

                \end{tcolorbox} 
            \end{minipage}
        \end{tcolorbox}
            \caption{\textbf{
    \ourenvsym{} User prompt, message and fact formats.}}
    \label{fig:sym_formats}
    \end{minipage}
\end{figure*}

\begin{figure*}[!ht]
    \centering
    \begin{tcolorbox}[colback=gray!10,arc=0pt,outer arc=0pt,boxrule=0pt,toprule=0.5mm,bottomrule=0.5mm,rightrule=0.5mm,leftrule=0.5mm] 
    \small
        \begin{minipage}{\linewidth} 
            \begin{tcolorbox}[colback=blue!5,
            colframe=blue!50,arc=0pt,outer arc=0pt,boxrule=0pt,toprule=0.5mm,bottomrule=0.5mm,rightrule=0.5mm,leftrule=0.5mm, title=Accuser System Prompt,coltitle=black]
                \textbf{Introduction:} \\
                - Your name is NAME, and you are a helpful AI assistant participating in a cooperative game with another player: \texttt{PARTNER\_NAME}. \\

                \textbf{Game Overview:} \\
                - The game involves \texttt{SUSPECT\_NUM} characters, identified by numbers 1 to \texttt{SUSPECT\_NUM}. One of the characters has been selected as the Winner and will be awarded a prize. \\
                - You and your partner have access to different information. You know the Winner's description, while your partner knows the description of every character but doesn't know which one is the Winner. \\
                - There is only one Winner, and only that character will exactly fit the Winner's description. \\
                - You will receive information in the following format: \\
                \begin{enumerate}
                    \item A list of properties about the Winner.
                    \item The current communication channel between you and your partner, containing shared properties.
                \end{enumerate}

                \textbf{Game Modes:} \\

                \textbf{1. Game Mode:} \\
                - The game proceeds in rounds. In each round, you can either award the prize to a character, making them the Winner, or ask your partner for information. \\
                - If there is only one character matching the criteria according to your information, you should award them the prize. Otherwise, ask your partner for the most helpful information. \\
                - You can ask for information in two ways: \\
                \begin{enumerate}
                    \item Ask if a specific character has a specific value for a property, and get a boolean answer. You must include both a property and a value for this action.
                    \item Ask for general info, receiving a property and value combination, and a list of all characters that fit that combination. Your partner will decide what property to share.
                \end{enumerate}

                - The goal is to find the correct Winner in a minimum number of rounds. Once you award a character, the game will end. \\
                - \textbf{Output Format in Game Mode:} \\
                \begin{verbatim}
{
 "thoughts": "Thoughts describing your optimal action at this point. 
 Describe whether to request information or award a character, and why you chose this action.",
 "action": 1, 2 or 3, // 1 to request specific information, 2 to request general information,
 and 3 to award a character
 "character": Character id to request information about or award, depending on your action. 
 This must be a number.,
 "property": "Relevant if action is 1. The property to ask for information about.",
 "value": "Relevant if action is 1. Value of the property that is asked about"
}
                \end{verbatim}

                \textbf{2. Analysis Mode:} \\
                - In this mode, you will answer questions about the task. Rounds do not count towards the goal, and you will not request or award. \\
                - A different output format will be provided. \\

                \textbf{Instructions Recap:} \\
                - Focus on minimizing rounds to find the Winner. \\
                - Award as soon as you are sure of who the Winner is. \\
                - Request informative facts to narrow down the possibilities. \\
                - Follow the specified JSON format for output in Game mode. \\
                - Use your knowledge to eliminate characters who cannot be the Winner and to narrow down the possibilities. \\

                Possible attributes for Characters are: \texttt{CHARACTER\_ATTRIBUTES} \\

                The Winner is described as: \\
                \texttt{WINNER\_DESCRIPTION}
            \end{tcolorbox} 
        \end{minipage}
    \end{tcolorbox}
    \caption{\textbf{
    \ourenvasym{} accuser system prompt.}}
    \label{fig:asym_sysprompts}
\end{figure*}

\begin{figure*}[!ht] 
    \centering
    \begin{tcolorbox}[colback=gray!10,arc=0pt,outer arc=0pt,boxrule=0pt,toprule=0.5mm,bottomrule=0.5mm,rightrule=0.5mm,leftrule=0.5mm] 
    \small
        \begin{minipage}{\linewidth} 
            \begin{tcolorbox}[colback=blue!5,
            colframe=blue!50,arc=0pt,outer arc=0pt,boxrule=0pt,toprule=0.5mm,bottomrule=0.5mm,rightrule=0.5mm,leftrule=0.5mm, title=Intel System Prompt,coltitle=black]
                \textbf{Introduction:} \\
                Your name is \texttt{NAME}, and you are a helpful AI assistant participating in a cooperative game with another player, \texttt{PARTNER\_NAME}. \\

                \textbf{Game Overview:} \\
                The game involves characters, with one character identified as the Winner. \\
                You have descriptions for all characters but do not know which one is the Winner. \\
                Beth has the description of the Winner, but not the character descriptions. \\
                You must help your partner discover which of the characters is the winner. \\
                You will receive information in the following format: \\
                \begin{enumerate}
                    \item A list of characters with their descriptions.
                    \item The current communication channel between you and Beth, containing shared properties.
                \end{enumerate}

                \textbf{Game Modes:} \\

                \textbf{1. Game Mode:} \\
                The game proceeds in rounds. In each round, you will receive a request from your partner. \\
                The request would either be a Boolean Question about one of the characters or a request for a Broad Message. \\
                Boolean question: Given the property, value and one possible character, answer whether that combination is true or false. \\
                Broad Message: Share the value of a single property and a list of characters that have that value for the property. \\
                Where sharing a broad message, you choose what property to share. Try sharing the most efficient property and value combination. \\
                The goal is to identify the Winner in the minimum number of rounds. \\

                \textbf{Output Format in Game Mode:} \\
                \begin{verbatim}
{
"thoughts": "Describe your optimal action at this point. 
Explain what is your response and why.",
"action": 1 or 2 // 1 to answer a request, 2 for a broad message,
"value": If the action is 1, this should be a boolean True/False value. 
If the action is 2, this should be a combination of a property name and a value separated by a -:
property-value
"character": "Relevant only if action is 2. 
List of all characters who have the property and value combination shared."
}
                \end{verbatim}

                \textbf{2. Analysis Mode:} \\
                In this mode, you will answer questions about the task. \\
                Rounds in this mode do not count towards the goal, and you will not make requests or accusations. \\
                A different output format will be provided in this mode. \\

                \textbf{Instructions Recap:} \\
                Focus on minimizing rounds to identify the Winner. \\
                Fulfill Beth's requests or broadcast your own messages to narrow down possibilities. \\
                Follow the specified JSON format for output in Game Mode. \\
                Possible attributes for Characters are: \texttt{CHARACTER\_ATTRIBUTES} \\

                Following is information about the Characters: \\
                \texttt{SUSPECT\_INFORMATION}
            \end{tcolorbox} 
        \end{minipage}
    \end{tcolorbox}
        \caption{\textbf{
    \ourenvasym{} Intel system prompts.}}
    \label{fig:asym_sysprompts_intel}
\end{figure*}

\begin{figure*}[!ht] 
    \centering
    \begin{minipage}{\linewidth} 
        \begin{tcolorbox}[colback=yellow!10,arc=0pt,outer arc=0pt,boxrule=0pt,toprule=0.5mm,bottomrule=0.5mm,rightrule=0.5mm,leftrule=0.5mm] 
            \begin{minipage}{\linewidth}
                \begin{tcolorbox}[colback=blue!5,
            colframe=blue!50,arc=0pt,outer arc=0pt,boxrule=0pt,toprule=0.5mm,bottomrule=0.5mm,rightrule=0.5mm,leftrule=0.5mm, title=User Prompt,coltitle=black]

                    The current communication channel between you and the other players is: \\
                    \texttt{COMMUNICATION\_CHANNEL} \\

                    The current turn is \texttt{TURN\_COUNT} out of \texttt{MAX\_TURN\_COUNT} \\
                    It’s your turn. What would you do?

                \end{tcolorbox} 
            \end{minipage}
        \end{tcolorbox}
                    \caption{\textbf{
    \ourenvasym{} User Prompt, which is identical between Accuser and Intel.}}
    \label{fig:asym_userprompts}
    \end{minipage}
\end{figure*}

\begin{figure*}[!ht] 
    \centering
    \begin{minipage}{\linewidth} 
        \begin{tcolorbox}[colback=gray!10,arc=0pt,outer arc=0pt,boxrule=0pt,toprule=0.5mm,bottomrule=0.5mm,rightrule=0.5mm,leftrule=0.5mm] 
            \begin{minipage}{\linewidth}
                \begin{tcolorbox}[colback=blue!5,
            colframe=blue!50,arc=0pt,outer arc=0pt,boxrule=0pt,toprule=0.5mm,bottomrule=0.5mm,rightrule=0.5mm,leftrule=0.5mm, title=Agent Messages,coltitle=black]

                    \begin{tcolorbox}[colback=white,arc=0pt,outer arc=0pt,boxrule=0.2mm,toprule=0.2mm,bottomrule=0.2mm,rightrule=0.2mm,leftrule=0.2mm, title=Request Specific]
                        Agent \texttt{NAME} has requested information: \texttt{BOOLEAN\_QUESTION}
                    \end{tcolorbox}

                    \begin{tcolorbox}[colback=white,arc=0pt,outer arc=0pt,boxrule=0.2mm,toprule=0.2mm,bottomrule=0.2mm,rightrule=0.2mm,leftrule=0.2mm, title=Request Broad]
                        Agent \texttt{NAME} has asked for general information (a broad message)
                    \end{tcolorbox}

                    \begin{tcolorbox}[colback=white,arc=0pt,outer arc=0pt,boxrule=0.2mm,toprule=0.2mm,bottomrule=0.2mm,rightrule=0.2mm,leftrule=0.2mm, title=Specific Answer]
                        Agent \texttt{NAME} has responded that character \texttt{CHARACTER} [is | is not] \texttt{ANSWER}
                    \end{tcolorbox}

                    \begin{tcolorbox}[colback=white,arc=0pt,outer arc=0pt,boxrule=0.2mm,toprule=0.2mm,bottomrule=0.2mm,rightrule=0.2mm,leftrule=0.2mm, title=Broad message]
                        Agent \texttt{NAME} has decided to return a broad message: For characters \texttt{CHARACTER\_LIST}, the property \texttt{PROPERTY} is \texttt{ANSWER}
                    \end{tcolorbox}

                \end{tcolorbox} 
            \end{minipage}
        \end{tcolorbox}
            \caption{\textbf{
    \ourenvasym{} message formats.}}
    \label{fig:asym_formats}
    \end{minipage}
\end{figure*}

\section{\ourenv{} Full Example - Hallucination}
\label{app:full_example}
For better understanding, we bring here the full text of the hallucination example, shown in \S\ref{sec:analysis}. Due to length, the example is split into 2 figures, the system prompt (Fig.~\ref{fig:ex_halluc_system}) and the user prompt and assistant answer (Fig.~\ref{fig:ex_halluc_user}).

\begin{figure*}[ht]
    \centering
    \begin{tcolorbox}[colback=gray!5,colframe=gray!50,title={Hallucination example - System prompt. Culprit is character 3}]
    \scriptsize
    \begin{verbatim}
System: 
**Introduction:**
- Your name is `Beth`, and you are a helpful AI assistant participating in a cooperative game with another player: Alex.

**Game Overview:**
- The game involves 10 characters, identified by numbers 1 to 10.
One of the characters has been selected as the Winner and will be awarded a prize.
- You and your partner have access to  different information.
   You know the Winner's description, while your partner knows the description of every character
   but doesn’t know which one is the Winner.
- There is only one Winner, and only that character will exactly fit the Winner’s description.
- You will receive information in the following format:
  1. A list of properties about the Winner.
  2. The current communication channel between you and your partner, containing shared properties.

**Game Modes:**

1. **Game Mode:**
- The game proceeds in rounds. In each round, you can either award the prize to a character,
making them the Winner, or ask your partner for information.
- If there is only one character matching the criteria according to your information, you should award them the prize.
  Otherwise, ask your partner for the most helpful information.
- You can ask for information in two ways:
  1. Ask if a specific character has a specific value for a property, and get a boolean answer.
    You must include both a property and a value for this action.
  2. Ask for general info, receiving a property and value combination, and a list of all characters that fit that combination.
    Your partner will decide what property to share.

- The goal is to find the correct Winner in a minimum number of rounds. Once you award a character, the game will end.
- **Output Format in Game Mode:**
```json
{
 "thoughts": "Thoughts describing your optimal action at this point.
   Describe whether to request information or award a character, and why you chose this action.",
 "action": 1, 2 or 3, // 1 to request specific information, 2 to request general information, and 3 to award a character
 "character": Character id to request information about or award, depending on your action. This must be a number.,
 "property": "Relevant if action is 1. The property to ask for information about.",
 "value": "Relevant if action is 1. Value of the property that is asked about"
}
```

**Instructions Recap:**
- Focus on minimizing rounds to find the Winner.
- Award as soon as you are sure of who the Winner is.
- Request informative facts to narrow down the possibilities.
- Follow the specified JSON format for output in Game mode.
- Use your knowledge to eliminate characters who cannot be the Winner and to narrow down the possibilities.
Possible attributes for Characters are: eye_color, eye_glasses, hair, hat, hobby, mood,
  pants, pants_color, shirt, shirt_color, shoe_color, watch.

The Winner is described as:
their eye color is blue, they have circular eye glasses, their hair is long, is wearing a brown hat, their hobby is basketball,
  is in a sad mood, they have red shoes, they have a bronze watch,
  they have short, black pants and a pink, button-up shirt.

The current communication channel between you and the other players is:
Agent Beth has requested information: is property eye_glasses of character 1 circular?
Agent Alex has responded that Character 1 is not wearing circular eye glasses
Agent Beth has asked for general information (a broad message)
Agent Alex has decided to return a broad message: For characters: [1, 3, 5, 8, 9], the property hobby is basketball
Agent Beth has requested information: is property eye_glasses of character 3 circular?
Agent Alex has responded that Character 3 is wearing circular eye glasses
Agent Beth has requested information: is property eye_color of character 3 blue?
Agent Alex has decided to return a broad message: For characters: [1, 2, 4, 5, 6, 7, 9, 10], the property eye_color is brown
Agent Beth has asked for general information (a broad message)
Agent Alex has decided to return a broad message: For characters: [1, 5, 8, 10], the property hat is black
Agent Beth has requested information: is property eye_color of character 3 blue?
Agent Alex has decided to return a broad message: For characters: [3, 8], the property eye_color is blue
Agent Beth has requested information: is property hat of character 3 brown?
Agent Alex has decided to return a broad message: For characters: [2, 3, 6, 10], the property eye_glasses is circular
Agent Beth has requested information: is property hat of character 3 brown?
Agent Alex has decided to return a broad message: For characters: [1, 2, 3, 4, 5, 6, 7, 9], the property shoe_color is red
Agent Beth has requested information: is property hat of character 3 brown?
Agent Alex has responded that Character 3 is wearing a brown hat
Agent Beth has requested information: is property hair of character 3 long?
Agent Alex has decided to return a broad message: For characters: [1, 4, 8, 9], the property mood is happy
Agent Beth has requested information: is property hair of character 3 long?
Agent Alex has responded that Character 3's hair is long
Agent Beth has requested information: is property mood of character 3 sad?
Agent Alex has responded that Character 3 is not in a sad mood

The current turn is 25 out of 31. It's your turn. What would you do?
    \end{verbatim}
    \end{tcolorbox}
    \caption{\textbf{Hallucination} example - System prompt.}
    \label{fig:ex_halluc_system}
\end{figure*}

\begin{figure*}[ht]
    \centering
    \begin{tcolorbox}[colback=gray!5,colframe=gray!50,title={Hallucination example - User prompt and answer. Culprit is character 3}]
    \scriptsize
The current communication channel between you and the other players is: \\
Agent Beth has requested information: is property eye\_glasses of character 1 circular? \\
Agent Alex has responded that Character 1 is not wearing circular eye glasses \\
Agent Beth has asked for general information (a broad message) \\
Agent Alex has decided to return a broad message: For characters: [1, 3, 5, 8, 9], the property hobby is basketball \\
Agent Beth has requested information: is property eye\_glasses of character 3 circular? \\
Agent Alex has responded that Character 3 is wearing circular eye glasses \\
Agent Beth has requested information: is property eye\_color of character 3 blue? \\
Agent Alex has decided to return a broad message: For characters: [1, 2, 4, 5, 6, 7, 9, 10], the property eye\_color is brown \\
Agent Beth has asked for general information (a broad message) \\
Agent Alex has decided to return a broad message: For characters: [1, 5, 8, 10], the property hat is black \\
Agent Beth has requested information: is property eye\_color of character 3 blue? \\
Agent Alex has decided to return a broad message: For characters: [3, 8], the property eye\_color is blue \\
Agent Beth has requested information: is property hat of character 3 brown? \\
Agent Alex has decided to return a broad message: For characters: [2, 3, 6, 10], the property eye\_glasses is circular \\
Agent Beth has requested information: is property hat of character 3 brown? \\
Agent Alex has decided to return a broad message: For characters: [1, 2, 3, 4, 5, 6, 7, 9], the property shoe\_color is red \\
Agent Beth has requested information: is property hat of character 3 brown? \\
Agent Alex has responded that Character 3 is wearing a brown hat \\
Agent Beth has requested information: is property hair of character 3 long? \\
Agent Alex has decided to return a broad message: For characters: [1, 4, 8, 9], the property mood is happy \\
Agent Beth has requested information: is property hair of character 3 long? \\
Agent Alex has responded that Character 3's hair is long \\
Agent Beth has requested information: is property mood of character 3 sad? \\
Agent Alex has responded that \textbf{Character 3 is not in a sad mood} \\

The current turn is 25 out of 31. \\
It's your turn. What would you do? \\

Answer:  '```json\{
"thoughts": "Character 3 matches most of the Winner\'s description, \textbf{but we found out they are not in a sad mood}.
We should now look for another character that matches the description of having blue eyes,
  circular glasses, long hair, a brown hat, and being in a sad mood.
Since we already know that character 3 has blue eyes, circular glasses, long hair,
  and a brown hat, we should look for another character that also has these properties \textbf{and is in a sad mood.}",
"action": 2,
"character": 0,
"property": "",
"value": ""
\}
```
I am asking for general information because we need to find another character that matches most of the Winner\'s description
  and is in a sad mood.
We can then ask more specific questions to narrow down the possibilities.'

    \end{tcolorbox}
    \caption{\textbf{Hallucination} example - User prompt and response.}
    \label{fig:ex_halluc_user}

\end{figure*}